\documentclass[pre, twocolumn,floatfix,superscriptaddress]{revtex4}

\usepackage{listings}
\usepackage{graphicx}
\usepackage{epstopdf}
\usepackage{multirow}
\usepackage{color}
\usepackage{amssymb,amsfonts,amsmath}
\usepackage[english]{babel}
\usepackage{float}

\begin{document}

\title{A high-reproducibility and high-accuracy method for automated topic classification}

\author
{Andrea Lancichinetti$^{1,2}$, M. Irmak Sirer$^{2}$, Jane X. Wang$^{3}$,
Daniel Acuna$^{4}$, Konrad K\"ording$^{4}$, Lu\'{\i}s A. Nunes Amaral$^{1,2,5,6}$\\
\normalsize{$^{1}$Howard Hughes Medical Institute (HHMI),}
\normalsize{$^{2}$ Department of Chemical and Biological Engineering,}
\normalsize{Northwestern University, Evanston, Illinois, USA,}
\normalsize{$^{3}$ Department of Medical Social Sciences} 
\normalsize{Northwestern University Feinberg School of Medicine, Chicago, Illinois, USA,}
\normalsize{$^{4}$ Department of Physical Medicine and Rehabilitation, Rehabilitation Institute of Chicago}
\normalsize{Northwestern University, Chicago, Illinois,  USA,}
\normalsize{$^{5}$ Northwestern Institute on Complex Systems, Northwestern University, Evanston, Illinois, USA,}
\normalsize{$^{6}$ Department of Physics and Astronomy, Northwestern University, Evanston, Illinois, USA.}
}

\begin{abstract}
Much of human knowledge sits in large databases of unstructured text.
Leveraging this knowledge requires algorithms that extract and record metadata
on unstructured text documents. 
Assigning topics to documents will enable intelligent search, statistical characterization, and meaningful classification. 
Latent Dirichlet allocation (LDA) is the state-of-the-art in topic classification. 
Here, we perform a systematic theoretical and numerical analysis 
that demonstrates that current optimization techniques for LDA
often yield results which are not accurate in inferring the 
most suitable model parameters.
Adapting approaches for community
detection in networks, we propose a new algorithm which displays high-reproducibility and
high-accuracy, and also has high computational efficiency.
We apply it to a large set of documents in the English Wikipedia 
and reveal its hierarchical structure.
Our algorithm promises to make ``big data'' text analysis systems more reliable.
\end{abstract}

\maketitle

The amount of data that we are currently collecting and storing is
unprecedented. A challenge for its analysis is that nearly $80\%$ of this data is in the form of unstructured text. 
As digital data keep increasing, there is a pressing need for fast and reliable algorithms 
to navigate and turn them into new knowledge.
One of the central challenges in the field of natural language processing 
is bridging the gap between information in text databases and their meaning 
in terms of topics. Topic classification algorithms are key for filling this gap.

Topic models use a database of text documents to automatically describe each document in terms of the underlying topics. This is the foundation for text recommendation systems \cite{Jin2005, Krestel2009}, digital image processing \cite{Sudderth05learninghierarchical, Niebles08unsupervisedlearning}, computational biology analysis \cite{Liu2010}, spam filtering \cite{Biro2008} and countless other modern-day technological applications.
Because of their importance, there has been an extraordinary amount of research and a number of different implementations of topic model algorithms\cite{deerwester1990indexing, lee1999learning, hofmann1999PLSA,
  blei2003LDA, steyvers2007probabilistic, hoffman2012sparse, anandkumar2012spectral, arora2013practical}.

\begin{figure*}
\begin{center}
\includegraphics[width=17.8cm]{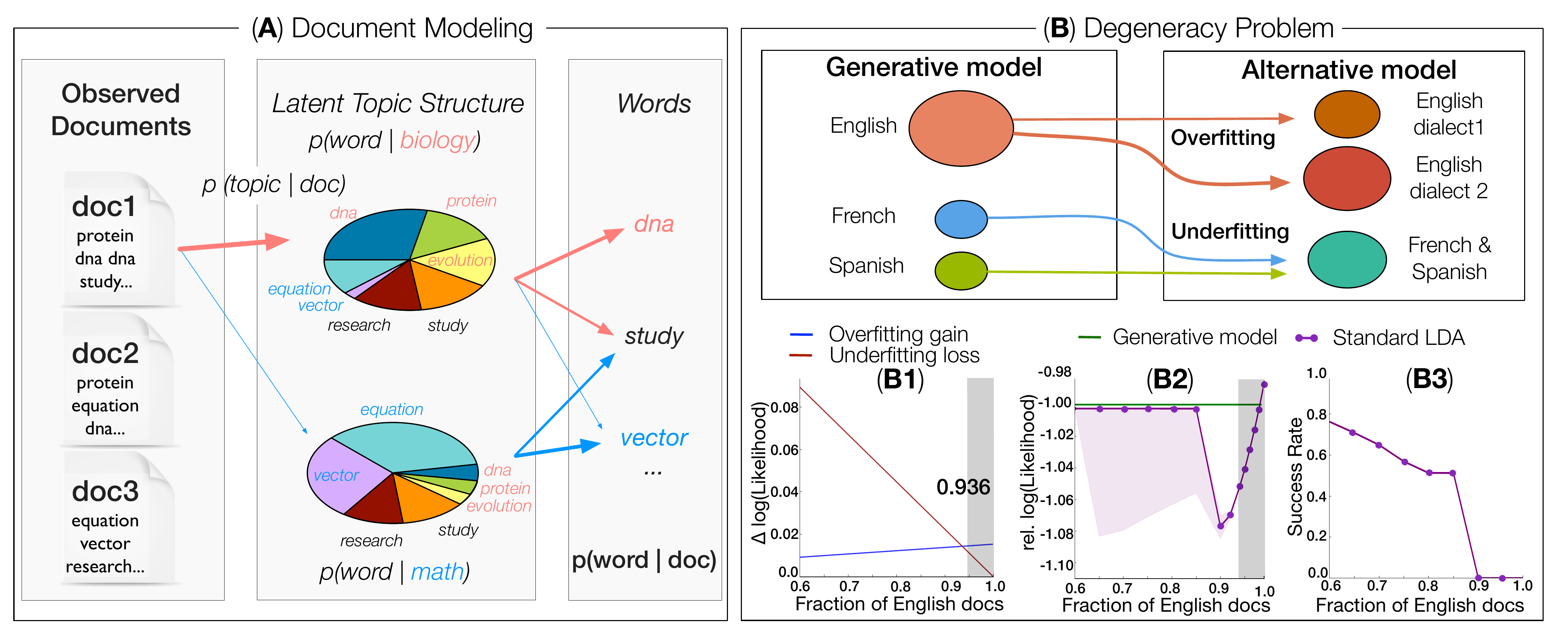}
\caption{
\textbf{A.} How documents are modeled as mixture of topics: 
we count the word frequencies in each document and we model these as mixtures of different topics, \textit{math} and \textit{biology} in this example. The topic structure is  \textit{latent}, meaning we do not have information about the ``true'' topics which generated the corpus. However, they can be estimated by a topic model algorithm.
\textbf{B}. 
We use data where each document is written in either English, French or Spanish (language = topic). However, a topic algorithm might separate English documents (many) into two subtopics, while French and Spanish (small groups) are merged.  
\textbf{B1}. We consider an example where each language has a vocabulary of 20 words, and the document length is 10 words: with this choice of the parameters, the alternative model has a better likelihood (for PLSA or symmetric LDA) if the fraction of English documents is bigger than $\sim 0.94$, when the overfitting gain compensates the underfitting loss. 
\textbf{B2}. LDA's performance (variational inference): the curve represents the median likelihood of the model inferred by the algorithm, while the shaded area delimits the $25$th and $75$th percentiles. The algorithm does not find the right generative model (green curve) before the theoretical limit (grey shaded area). \textbf{B3}.  Probability that LDA infers the actual generative model. 
}
\label{fig::intro}
\end{center}
\end{figure*}

Latent Dirichlet allocation (LDA)
\cite{blei2003LDA,blei2003hierarchical,blei2007correlated} is the
state-of-the-art method in topic modeling.  As its predecessor,
Probabilistic latent semantic analysis (PLSA) \cite{hofmann1999PLSA},
it relies on fitting a generative model of the \textit{corpus}
(Fig.~\ref{fig::intro}). Specifically, the model assumes that a document
$doc$ in the corpus covers a mixture of topics, and that each topic is
characterized by a specific word usage probability distribution.  For instance,
consider a corpus of documents addressing two topics, mathematics and
biology. Each document in the corpus will cover these topics with
given probabilities.
A biology focused document d$_{bio}$, for
example, might have $p(\textrm{biology}|\textrm{d}_{bio})=90\%$ and
$p(\textrm{math}|\textrm{d}_{bio})=10\%$.
Documents focused on different topics will make use of different words.
Some words will be used for biology documents
such as \textit{dna} or
\textit{protein} because $p(\textrm{dna}| \textrm{biology}) \gg
p(\textrm{dna}| \textrm{math})$. In contrast, words such as
\textit{tensor} or \textit{equation} will primarily be used in a
math-focused document because $p(\textrm{tensor}| \textrm{biology})
\ll p(\textrm{tensor}| \textrm{math})$.  Additionally, there will be
words such as \textit{research} or \textit{study} that are generic and 
will be
used equally by both topics.

In practical applications, one has access to the word counts in each document, 
but the topic structure will be unobservable or 
\textit{latent}. The challenge thus is to estimate the topic structure which is 
defined by the probabilities $p(topic|doc)$ and $p(word|topic)$.  PLSA and LDA both try
to estimate the model with the highest probability of generating the data
\cite{blei2003LDA, Griffiths2004finding, Nallapati2007, hofmann1999PLSA}, but PLSA does
not account for the probability of choosing a certain topic
mixture. Crucially, both methods rely on maximization of a likelihood
that depends non-linearly on a large number 
of variables, 
a NP-hard problem \cite{sontag2011complexity}.

Although it is well known that the problem is computationally hard, 
little is known about how in practice the roughness of the likelihood landscape
impacts an algorithm's performance.
In order to gain a more thorough theoretical understanding,
we implement a controlled analysis
of a highly specified and constructed set of data.
This high degree of control allows us to tease apart the
theoretical limitations of the algorithms from other sources 
of error that would be normally uncontrolled in traditional datasets.
Our analysis reveals that standard techniques for likelihood optimization 
are hindered by the very rough topology of the 
landscape, even in very simple cases such as when topics use exclusive vocabularies. 
We show that a network approach 
 to topic modeling enables searching the likelihood landscape much more efficiently, 
 yielding more accurate and reproducible results.

\section{The likelihood landscape is always rough}

Most practitioners know that a very large number of topic models
can fit the same data almost equally well: this poses a serious problem
for an algorithm's stability. 
We start investigating this problem by considering
an elementary test-case, which we denote the
language corpus.
``Toy'' models are helpful because they can be analytically treated
and provide useful insights 
for more realistic and complex cases.

In our language corpus, topics are
fully disambiguated languages -- that is, no similar words are used
across languages -- and each document is written entirely in a single
language, thus creating the simplest possible test case.
As is assumed by the LDA generative model, we use a two-step
process to create synthetic documents.  In the first step, we select a
language with probability $p(language)$,  which corresponds to a Dirichlet distribution
 with very small concentration parameters (see SI).
Given the language, in the
second step, we randomly sample a given number of words from that
language's vocabulary into the document. For the sake of simplicity,
we restrict the vocabulary of each language to a set of $N_w$ unique
equiprobable words.  Thus, an ``English'' document in the language
corpus, is just a ``bag'' of English words.
Note that every document uses words from a single language.

Let us be more concrete and consider a dataset with three languages
and distinct number of documents in each language.
Consider also that there are more documents in English
than in the other languages.
An implementation of a topic model algorithm could correctly 
infer the three languages as topics or alternatively split ``English'' 
into two ``dialects'' and merge two other languages (see Fig.~\ref{fig::intro}).  This alternative
model is wrong on two counts: it splits English into two parts, while merging
two different languages.
Na\"{\i}vely, one would expect the alternative model
to have a smaller likelihood than the correct generative model.
However, this is not always the case  (Fig.~\ref{fig::intro}C), for PLSA \cite{hofmann1999PLSA}
and the symmetric version of LDA
 \footnote{Symmetric LDA implements a model where each topic 
  can be chosen a priori with equal probability.}
\cite{blei2003LDA,
  Griffiths2004finding}. In fact, dividing (or overfitting) the
``English'' documents in the corpus yields an increase of the
likelihood. As we show in the SI text, the log-likelihood increases by between
$\log^22$ and $1/\pi$ per English document, depending on the average
length of the documents, through this process of overfitting. Analogously,
merging (or underfitting) the ``French'' and ``Spanish'' documents, 
results in a decrease of the log-likelihood of $L_d \log 2$
per ``French'' and ``Spanish'' document, where $L_d$ is the average
length of the documents. Thus, there is a critical fraction of ``French'' and
``Spanish'' documents below which the alternative model will
have a greater likelihood that the correct generative model (Fig.~\ref{fig::intro}).

Note that this theoretical limit of
a likelihood's ability of identify the correct
generative model is not limited to topic modeling. Indeed, it also holds
for non-negative matrix
factorization
\footnote{
Non-negative matrix factorization is a popular
low-rank matrix approximation algorithm which has found
countless applications, for example, in
face recognition, text mining and high-dimensional data analysis
in general. 
}
\cite{lee1999learning} with KL-divergence, because of
its equivalence to PLSA \cite{gaussier2005relation}.

However, the critical size of underfitted documents
depends on the length $L_d$ of the documents in the corpus,
and decreases as $1/L_d$. In fact, increasing the documents' length 
or using asymmetric LDA
\cite{wallach2009rethinking} rather than symmetric LDA, one can
show, for the language corpus, that the generative model always
has a higher likelihood than the alternative model (see SI). In this
case, the ratio of the log-likelihood of the alternative model and the
generative model can be expressed as,
\begin{equation}
\frac{\langle \log {\mathcal{L}_{\textrm{alt}} } \rangle} { \langle \log {\mathcal{L}_{\textrm{true}}} \rangle} \simeq 1  -  \frac{ f_{U}  \log2 }{\log N_w}.
\end{equation}
\noindent
where $\mathcal{L}_{\textrm{alt}}$ and $\mathcal{L}_{\textrm{true}}$
are the likelihoods of the alternative and generative model
respectively and $f_{U}$ is the fraction of underfitted documents
(``French'' and ``Spanish,'' in the example).

Even though the generative model has a greater likelihood, the ratio
on the left-hand side of Eq.~(1) can be arbitrarily close to 1.  The
reason is that the ratio is independent of the number of documents in
the corpus and of the length of the documents.  Thus, even
with an infinite number of infinitely long documents, the generative
model does not ``tower'' above other models in the likelihood
landscape.  The consequences of this fact are important
because the number of alternative latent models that can be
defined is extremely large -- with a vocabulary of 1000 words per
language, the number of alternative models is on
the order of $10^{300}$
(see SI).

In conclusion, we find that only the full Bayesian model \cite{wallach2009rethinking}
can potentially detect the correct generative model regardless of the documents' length, 
and an extremely large number of models are very close to the correct one in terms of likelihood.
In the next section, we will show how current optimization techniques are affected by this problem.

\section{Numerical analysis of the language test}

Although the language corpus is a highly idealized case, 
it provides an example where many competing models
have very similar likelihood and the overwhelming majority of those models
have more equally sized topics. 
Indeed, because of the high degeneracy of the likelihood landscape,
standard optimization techniques might not find 
the model with highest likelihood even in such simple cases,  and 
they might yield different models 
across different runs,
as it has been previously reported
 \cite{wallach2009rethinking,
  steyvers2007probabilistic}.

\begin{figure}
\begin{center}
\includegraphics[width=8.7cm]{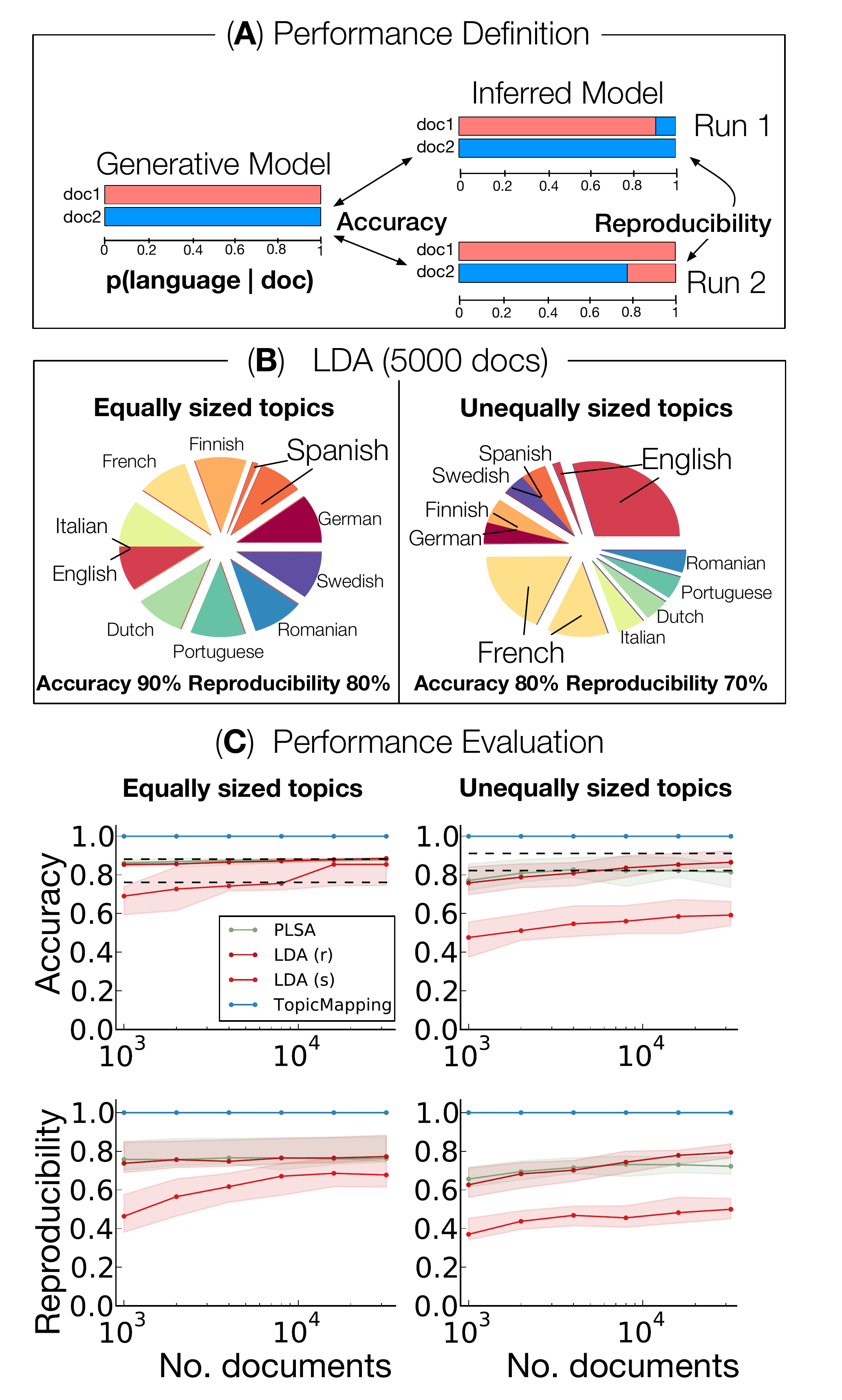}
\caption{
Performance of the algorithms on a datasets of documents written in different languages.
To write each document, we first decide which language to use, and we then write 100 words sampled with the actual word frequencies of that language (we use a limited vocabulary of the 1000 most frequent words). 
For the sake of simplicity all words have been disambiguated. 
\textbf{A}. The accuracy is measured in terms of the Best Match similarity (see Methods) among the fitted model and the generative model. Reproducibility is the similarity among fitted models obtained in different runs.
 \textbf{B}. The pie charts show the topics typically found by LDA standard optimization. Each slice is a topic found by the algorithm and it is divided in colored strips. Each color represents a different language. The area of the strips in each slice is proportional to the language probability in that topic.
\textbf{C}. Reproducibility and accuracy for this test tuning the number of documents (we show median values and 25th and 75th percentiles). We input the correct number of topic in LDA and PLSA. The dashed lines indicate the accuracy we would obtain overfitting one  language and underfitting two other languages (top line), or overfitting two languages and underfitting four (bottom line).
LDA(r) and LDA(s) refers to different initializations for the optimization technique (random or seeded).
}
\label{fig::languages}
\end{center}
\end{figure}

Moreover, since small topics are the hardest to resolve (see SI, Sec.~1.6),
standard algorithms might require the
assumption that there are more topics in the corpus than
in reality because the ``extra topics'' are needed to resolve
small topics.

We test these hypotheses numerically, on two synthetic language
corpora (Fig.~\ref{fig::languages}). For the first corpus, which we
denote egalitarian, each of ten languages comprises an equal number of documents.
For the second corpus, which we denote
oligarchic, 60\% of the documents belong to 20\% of the
languages. Specifically, we group the languages into two classes. The
first class comprises two languages with 30\% of the documents in the
corpus.  The second class comprises eight languages with 5\%
of the documents. For both corpora,
we used the real-world word frequencies \cite{InvokeIT} of the
languages.

In order to determine the validity of the models inferred by the
algorithm under study, we calculated both the accuracy and the
reproducibility of the algorithms' outputs.  We use a measure
of normalized similarity (see Methods) to compare the inferred model to the
generative model (accuracy) and to compare the inferred models from
two runs of the algorithm (reproducibility).

In the synthetic corpora that we consider,
topics are not unequal enough and documents are sufficiently long,
so that both datasets have their highest likelihood for the generative model,
and for PLSA and symmetric LDA.  
Additionally, we run
the standard algorithms \cite{hofmann1999PLSA, blei2003LDA} with the number of topics in the generative
model (as we show in the SI, estimating the number of topics via model
selection would lead to an over-estimation of the number of
topics). We find that PLSA and 
the standard optimization algorithm implemented with
LDA (variational inference) \cite{blei2003LDA} are unable to find the global maximum
of the likelihood landscape (see Fig.~\ref{fig::languages}).
In the SI we also show the results for asymmetric LDA implementing Gibbs sampling \cite{wallach2009rethinking},
which, interestingly, performs well only in the egalitarian case.

Our results thus show that it is highly
inefficient to explore the likelihood landscape blindly, either by starting from
random initial conditions or by randomly seeding the topics using a sample of
documents (Fig.~\ref{fig::languages}), as is the current standard practice.

\section{A network approach}

In order to improve on the performance of current methods, we surmise
that it will be useful to build some intuition about where to search
in the likelihood landscape.  We start by noting that a
corpus can be viewed as a bipartite network of words and
documents \cite{dhillon2001co}, and, using this insight, we construct a
network of words which are connected if they co-appear in a
document \cite{zhou2007bipartite}. 

In the language corpora, finding the languages is as simple
as finding the connected components of this graph.  In general,
however, finding topics will be more complex because of words shared
by topics. We propose a new approach comprising three
steps, which we denote TopicMapping.  In the first step, we filter out
words that are unlikely to provide a separation between topics because
they are used indiscriminately across topics.  Specifically, we
compare the dot-product similarity \cite{Tan2005IDM} of each pair of
words (which co-appear in at least one document) with the expectation
for a null-model where words are randomly shuffled across
documents. For the null-model, the distribution of dot-product
similarities of pairs of words is well approximated by a Poisson
distribution whose average depends on the frequencies of the words
(see SI).  We set a $p$-value of $5\%$ for accepting the significance
of the similarity between pairs of words.

In the second step, we cluster the filtered network of words 
using a clustering algorithm developed by 
Rosvall and Bergstrom (Infomap) \cite{rosvall2008maps}. 
Unlike standard topic modeling algorithms, the method does not
require an estimate of the number of topics present in the
corpus. We use the groups identified by the clustering
algorithm as our initial guesses for the number and word composition
of the topics.  Because our clustering algorithm is exclusive -- that
is, words can belong to a single topic -- we must use a latent topic
model which allows for non-exclusivity.  Specifically, we locally optimize a
PLSA-like likelihood in order to obtain our estimate of model
probabilities (see SI for more information).

In the third step, we can decide to refine our guess further running asymmetric
LDA likelihood optimization \cite{blei2003LDA}
using, as initial conditions, the model probabilities found in the previous step.
In general, if the topics are not too heterogeneously distributed,
the algorithm converges after only a few iterations, as our guess is generally 
very close to a likelihood maximum (we actually found only one case where more iterations were needed:
the Wikipedia dataset, see Fig.~\ref{fig::wikimain}).
Figure~\ref{fig::languages} shows the
excellent performance of the TopicMapping algorithm.

\section{A real world example}

In order to test the validity of the TopicMapping algorithm and better
compare its performance to standard LDA optimization methods,
we next consider a real-world
corpus comprising 23,838 documents obtained from Web of
Science (WoS).  Each document contains the title and the abstract of a paper
published in one of six top journals from different disciplines
(Geology, Astronomy, Mathematics, Biology, Psychology, Economics).
We pre-processed the documents in the WoS corpus by 
using a stemming algorithm \cite{porter2}
and removing
a standard list of stop-words. Pre-processing yielded 106,143
unique words.

\begin{figure}[h!]
\begin{center}
\includegraphics[width=8.7cm]{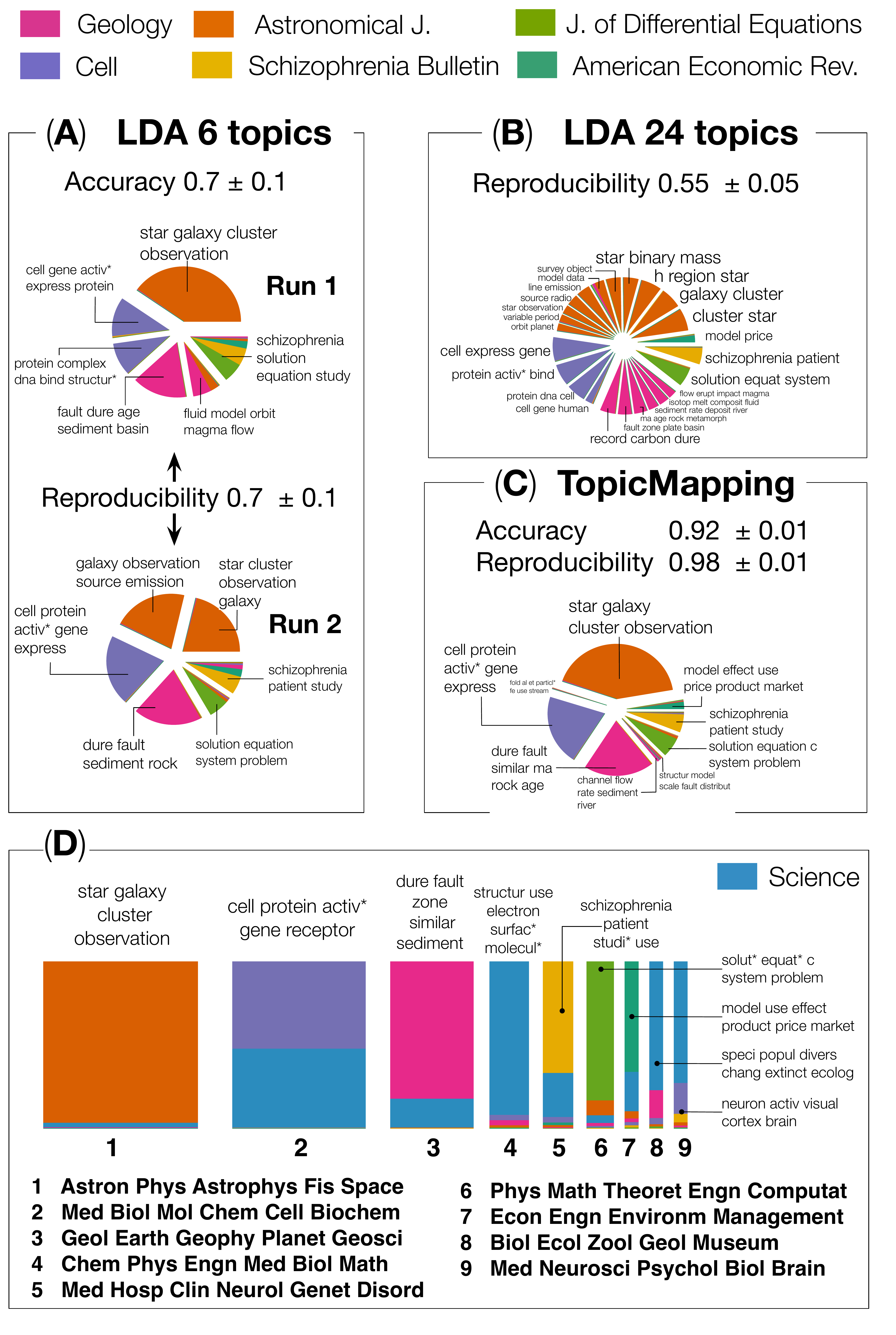}
\caption{
Performance of the algorithms on a real world example. In the pie charts, each slice is a different topic found by the method and the colored areas are proportional to the probability of the corresponding journal given that topic: $p(journal|topic)=\sum_{doc} p(journal|doc) \times p(doc|topic)$. The topic labels are the most frequent words in the topic. The ``*'' symbol is due to the stemming algorithm we used (porter2).
\textbf{A}. Performance of standard LDA when we input the number of journals as number of topics. Big topics are split and small ones are merged. \textbf{B}. Performance of LDA when we input the number of topics suggested by model selection. Small topics are now resolved but big ones are split so that each topic is comparable in size. \textbf{C.} TopicMapping's performance.  \textbf{D.} Topics found by TopicMapping in a corpus were we added an interdisciplinary journal such as Science. We also show the most frequent affiliations of papers published in Science in each topic (bottom). The total number of topics found is 19 but only topic with probability bigger than $2\%$ are shown in the figure (9 topics). }
\label{fig::wos}
\end{center}
\end{figure}

We surmised a generative model in which each journal defines a topic
and in which each document is assigned exclusively to the topic
defined by the journal in which it was published. We then compare the
topics inferred by symmetric LDA (variational inference) and TopicMapping with the
surmised generative model (Fig.~3). While TopicMapping has
nearly perfect accuracy and reproducibility, standard LDA optimization has a
significantly poorer performance.
When using the standard approach,
LDA estimates that the corpus comprises 20 to 30 topics (see
SI) and yields a
reproducibility of only 55\%.  Even when letting LDA know that there
are only six topics, the inferred models will put together papers from
small journals yielding an accuracy and a reproducibility of 70\%.

Adding an interdisciplinary journal (Science), 
we can see that TopicMapping assigns the majority of papers
published in Science to the already found topics, 
but several new topics are identified.
In terms of likelihood, TopicMapping yields a slightly better likelihood
than standard LDA optimization, but only if we compare models
with the same effective number of topics.
A more detailed discussion
on this point can be found in the SI.

\section{Systematic analysis on synthetic data} 

As a final and more systematic evaluation of the accuracy and
reproducibility of the different algorithms, we implement a
comprehensive generative model, where documents choose a topic
distribution from a Dirichlet distribution as proposed in the LDA model. We tune
the difficulty in separating topics within the corpora by
setting (1) the value of a parameter $\alpha$ which determines both
the extent to which documents mix topics, and the extent to which
words are significantly used by different topics; and (2) the
fraction of words which are generic, that is, contain no information
about the topics (see Methods).

\begin{figure}[ht!]
\begin{center}
\includegraphics[width=8.7cm]{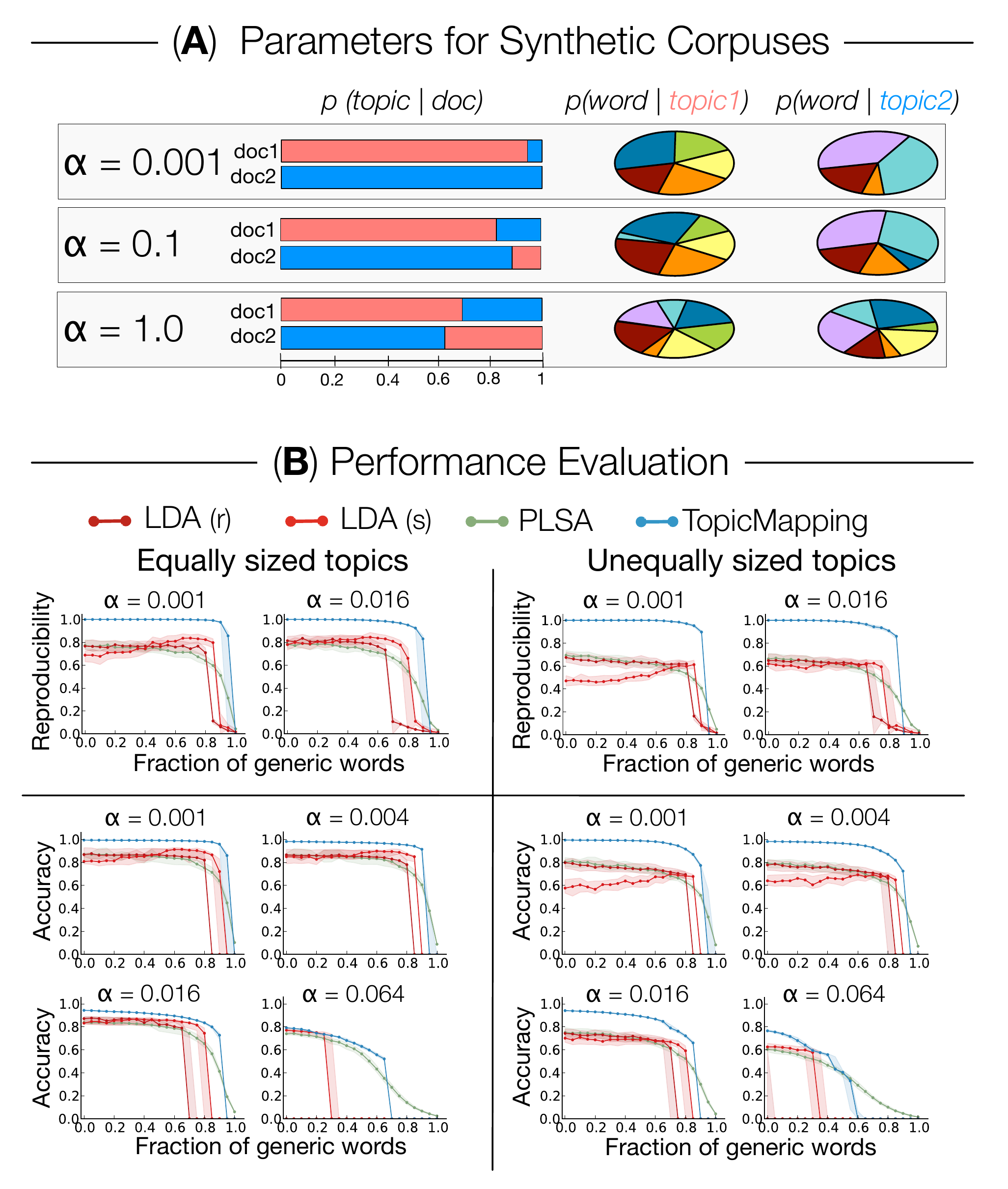}
\caption{
\textbf{A.} Creating synthetic corpora using the generative model. 
For each document, $p(topic | doc)$ is sampled from a Dirichlet distribution whose hyper parameters are defined as: $\alpha_{topic} = K \times p(topic) \times \alpha$, where $K$ is the number of topics, $p(topic)$ is the probability (i.e. the size) of \textit{topic} and $\alpha$ is a parameter which tunes how mixed documents are: smaller values of $\alpha$ yield a simpler model where documents make use of fewer topics. We also have a parameter to fix the fraction of generic words, and we implement a similar method for deciding $p(word | topic)$ for specific and generic words (see Methods). Once the latent topic structure is chosen, we write a corpus drawing words with probabilities given by the mixture of topics.
\textbf{B.}
The performance of the topic modeling algorithms on synthetic corpora. In all our tests, we generate a corpus of $1000$ documents, of $50$ words each, and our vocabulary is made of $2000$ unique equiprobable words. We set the number of topics $K=20$ and we input this number in LDA and PLSA.
``Equally sized'' means all the topics have equal probability $p(topic)=5\%$, while in the ``unequally sized'' case, 4 large topics have probability $15\%$ each, while the other 16 topics have probability $2.5\%$. LDA(s) and LDA(r)  refer to seeded and random initialization for LDA (variational inference). The plots show the median values as well as the 25th and 75th percentiles.
}
\label{fig::systematic}
\end{center}
\end{figure}

Fig.~\ref{fig::systematic} shows our results for the synthetic corpora.
We have also
done a more systematic analysis (see SI), but the main conclusion is
the same as for the language test: the generative model has the
highest likelihood (topics are sufficiently equal in size), but the number
of overfitting models is so large and they are so close in terms of
likelihood, that the optimization technique requires help in
exploring the right portion of the parameter space. Without the right
initialization, we get lower accuracy and reproducibility, as well as
equally sized topics and an overestimation of the number of topics (see SI).

\begin{figure*}[ht!]
\begin{center}
\includegraphics[width=17.8cm]{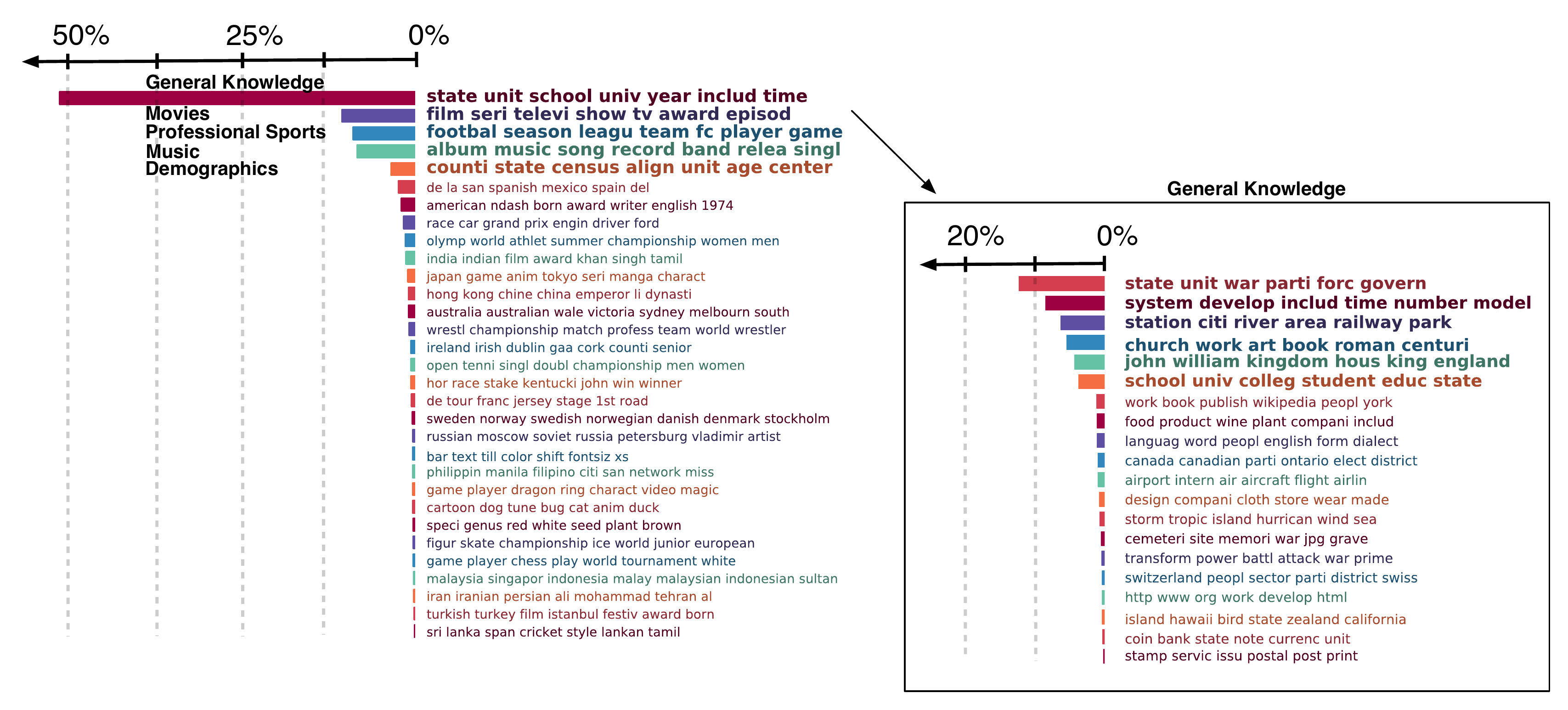}
\caption{Topics found by TopicMapping on a large sample of Wikipedia ($\sim 1.3$ million documents). Here, we show the topics found by the TopicMapping after one single LDA iteration: indeed, this dataset represents an example where optimizing LDA until convergence gives rather different results (see SI, Sec.~1.6 and Sec.~10). We highlight the top topics that account for the $80\%$ of total  documents: those are just a handful of topics which are very easy to interpret (left).  The inset shows the topics we find on the sub-corpus of documents assigned to the main topic ``General Knowledge''. }
\label{fig::wikimain}
\end{center}
\end{figure*}

The computational overhead of using TopicMapping, 
for obtaining an initial guess of the parameter values, 
is small and the 
algorithm can be easily parallelized. To demonstrate this fact, we applied 
TopicMapping 
to a sample of the English Wikipedia with more than a million documents
and almost a billion words (see Fig.~\ref{fig::wikimain}).

\section{Conclusions}

Ten years since its introduction,
there has been surprisingly little research on the limitations of
LDA optimization techniques for inferring topic
models \cite{wallach2009rethinking}. 
We are able to obtain a remarkable improvement in method validity by
using a much simpler objective function \cite{rosvall2008maps} to
obtain an educated guess of the parameter values in the latent
generative model. This guess is obtained exclusively using word-word
correlations to estimate topics, whereas word document correlations are accounted for
later in refining
the initial guess. The algorithm is related to some recent work on spectral algorithms
\cite{anandkumar2012spectral, arora2013practical}.
However, here we propose a practical implementation
which makes no assumption about topic separability or the number of topics, as most spectral algorithms do.
Interestingly, TopicMapping provides only  slight improvements
in terms of likelihood 
(because of the high degeneracy of the likelihood landscape),
but nevertheless yields much better accuracy and reproducibility.

\section{Methods}

\subsection{Comparing models}
\label{sec_compare::sec}
Here, we describe the algorithm for measuring the similarity between two models, $p$ and $q$.
Both topic models are described by two probability distributions: $p(topic|doc)$ and $p(word|topic)$.
Given a document, we would like to compare two distributions: $p(t'|doc)$ and $q(t''|doc)$. The problem is not trivial because the topics are not labeled: the numbers we use to identify the topics in each model are just one of the $K!$ possible permutations of their labels.
Instead, documents have of course  the same labels. For this reason, it is easy to quantify the similarity of topics  $t'$ and $t''$ from different models, if we look at which documents are in these topics: we can use Bayes' theorem to compute  $p(doc|t')$ and $q(doc|t'')$ and compare these two probability distributions. We propose to measure the distance between $p(doc|t')$ and $q(doc|t'')$ as the $1-$norm (or  Manhattan distance):
$\| p(doc|t') - q(doc|t'') \|_1 = \sum_{doc} |p(doc|t') - q(doc|t'')|$.
Since we are dealing with probability distributions, $\| p - q \|_1 \leqslant 2 $. We can then define the normalized similarity between topics $t'$ and $t''$ as: 
$s(t', t'')=1 -  \frac{1}{2}\| p(doc|t') - q(doc|t'') \|_1$.

To get a global measure of how similar one model is with respect to the other, we compare each topic $t'$ with all topics $t''$ and we pick the topic which is most similar to $t'$: this is the similarity we get best matching model $p$ versus $q$:
$\textrm{BM}(p \rightarrow q)=    \sum_{t'} p(t')  \, \max_{t''} s(t', t'')$,
where $\textrm{BM}$ stands for Best Match, and the arrow indicates that each topic in $p$ looks for the best matching topic in $q$.  
Of course, we can make this similarity symmetric,  averaging the measure with $\textrm{BM}( p \leftarrow q) =  \sum_{t''} q(t'')  \, \max_{t'} s(t', t'')  $ :
$\textrm{BM}(p,q)=  \frac{1}{2} \big [ \textrm{BM}(p \rightarrow q) + \textrm{BM}( p \leftarrow q) \big ]$.

Although this similarity is normalized between 0 and 1, it does not inform us about how similar the two models are compared to what we could get with random topic assignments. For this reason, we also compute the average similarity $\textrm{BM}(p \rightarrow q_s)$, where we randomly shuffle the document labels in model $q$. Our null model similarity is then defined as $\textrm{BM}_{rand}= \frac{1}{2} [ \textrm{BM}(p \rightarrow q_s)+\textrm{BM}(p_s \leftarrow q) ] $.

Eventually, we can define our measure of normalized similarity between the two models as:

\begin{equation}
\textrm{BM}_n= \frac{\textrm{BM} - \textrm{BM}_{rand}}{1-\textrm{BM}_{rand}} . 
\end{equation}

An analogous similarity score can be defined for words using $p(word|topic)$ instead of $p(doc|topic)$.

\subsection{Generating synthetic corpora}
\label{syn_section::sec}
The algorithm we used to generate synthetic datasets relies on the generative model assumed by LDA.
First, we specify the number of documents and the number of words in each document, $L_d$. For simplicity, we set the same number of words for each document, $L_d=L$.
Next, we set the number of topics $K$ and the probability distribution of each topic, $p(topic)$.
Finally, we specify the number of words in our vocabulary, $N_w$, and the probability distribution of each word, $p(word)$. For the sake of simplicity, we used uniform probabilities for $p(word)$, although the same model can be used for arbitrary probability distributions.
All these parameters define the size of the corpus, the other aspect to consider is how mixed documents are across topics and topics are across words: this can be specified by one hyper-parameter $\alpha$, whose use will be made clear in the following.
The algorithm works in the following steps:

\begin{enumerate}
  \item For each document $doc$, we decide the probability this document will make use of each topic: $p(topic | doc)$. These probabilities are sampled from the Dirichlet distribution with parameters: $\alpha_{topic} = K \times p(topic) \times \alpha$. The definition is such that $topic$ will be used in the overall corpus with probability $p(topic)$, while the factor $K$ is a normalization which assures that we get $\alpha_{topic} = \alpha$ for equiprobable topics. In this particular case, $\alpha=1$ means that documents are assigned to topics drawing the probabilities uniformly at random (see SI  for more on the Dirichlet distribution).
  \item For each topic, we need to define a probability distribution over words: $p(word|topic)$. For this purpose, we first compute $p(topic|word)$ for each word, sampling the same Dirichlet distribution as before ($\alpha_{topic} = K \times p(topic) \times \alpha$). Second, we get $p(word|topic)$ from Bayes' theorem: $p(word|topic) \propto p(topic|word) \times p(word)$.
  \item We now have all we need to generate the corpus. Every $word$ in document $doc$ can be drawn, first, selecting $topic$ with probability $p(topic|doc)$ and, second, choosing $word$ with probability $p(word|topic)$.  
\end{enumerate}

Small values of the parameter $\alpha$ will yield ``easy'' corpora where documents are mostly about one single topic and words are specific to a single topic, (Fig.~4). For simplicity, we keep $\alpha$ constant for all documents and words. However, it is highly unrealistic that all words are mostly used in a single topic, since every realistic corpus contains generic words. To account for this, we divide the words into two classes, specific and generic words: for the former class, we use the same $\alpha$ as above, while for generic words we set $\alpha=1$. The fraction of generic word is a second parameter we set.

\section{Supplementary Materials}
TopicMapping software, datasets and related codes are available at \url{https://sites.google.com/site/andrealancichinetti/topicmodeling}

\begin{acknowledgments}
We thank Xiaohan Zeng, David Mertens and Adam Hockenberry for discussions.
\end{acknowledgments}

\renewcommand{\figurename}{Figure}
\renewcommand{\thefigure}{S\arabic{figure}}
\renewcommand{\tablename}{Table}
\renewcommand{\thetable}{S\arabic{table}}
\renewcommand{\theequation}{S\arabic{equation}}

\renewcommand{\thesection}{S\arabic{section}}

\setcounter{figure}{0}
\setcounter{table}{0}
\setcounter{section}{0}

\bigskip \textbf{\LARGE Supplementary Information}\\ \bigskip

\section*{OUTLINE}

The supplementary material is organized as follows:

\begin{itemize}

\item Sec.~\ref{sec_deg} provides analytical insights on the likelihood landscape: in particular,  we discuss the theoretical limitations of PLSA \cite{hofmann1999PLSA} and symmetric LDA \cite{blei2003LDA} in finding the correct generative model. Also, in Sec.~\ref{seb_sec::modelcomp} we present an additional example which suggests why equally sized topics often have better likelihood.

\item Sec.~\ref{network_sec} describes the network approach we take for topic modeling.

\item Sec.~\ref{heldout_likelihood} shows that standard LDA tends to over-estimate the number of topics, and to find equally sized topics.

\item Sec.~\ref{additional_sec} presents a more detailed analysis of the synthetic datasets: among other things, we visualize the algorithms' results. 

\item Sec.~\ref{asym_LDA} shows the performance of asymmetric LDA \cite{wallach2009rethinking}.

\item Sec.~\ref{wos_hierarchy} discusses the hierarchical topics of Web of Science dataset and the role of the $p$-value for TopicMapping.

\item Sec.~\ref{sec_time} presents the computation complexity of TopicMapping.

\item Sec.~\ref{wikipedia} shows the topics we found on a large sample of the English Wikipedia.

\item Sec.~\ref{likelihood::sec} shows that TopicMapping often provides models with higher likelihood, if we compare models with the same effective number of topics.

\item Sec.~\ref{app_secs} is an appendix with some more technical information about the calculations presented in Sec.~\ref{sec_deg}, some clarifications about Dirichlet distributions and measuring perplexity, and some
technical information about the algorithms' usage.

\end{itemize}

\section{Degeneracy problem in inferring the latent topic structure}
\label{sec_deg}

\subsection{Introduction}

Most topic model optimizations are known to be computationally hard problems \cite{sontag2011complexity}. However, not much is known about how the roughness of the likelihood landscape affects the algorithms' performance.

We investigate this question by ($i$) defining a simple generative model, ($ii$) generating synthetic data accordingly  and ($iii$) measuring how well the algorithms recover the generative model (which is considered the ``ground truth").

In the whole study, we examine different generative models. In this section, we study the simplest among those, the language test.
For this model, we prove that, \textit{if the topics are not enough equally sized, the model which maximizes the likelihood optimized by} PLSA \textit{and symmetric} LDA \textit{can be different from the generative model}. More specifically, we show that it is possible to find an extremely large number of alternative models (with the same number of topics) which overfit some topics and underfit some others but have a better likelihood than the true generative model.
Symmetric LDA is the version of LDA where the prior $\alpha$ is assumed to be the same for all topics, and it is probably the most commonly used.
For asymmetric LDA, which allows different priors, the correct generative model has the highest likelihood, in the language test. However, we show that the ratio between the log-likelihood of the generative model and the one of the alternative models can be arbitrarily close to 1, even in the limit of infinite number of documents and infinite number of words per document.
 This implies that even increasing the amount of available information, the likelihood of the generative model will not increase relatively to the others. Below, we also give some quantitative estimates.

\subsection{The simplest generative model}

Let us call $K$ the number of topics. 
Each topic has a vocabulary of $N_w$ words, and for the sake of simplicity we assume all the words are equiprobable. We also assume that we cannot find the same word in two different topics, so that we are actually dealing with fully disambiguated languages. 
Then, each document is entirely written in one of the languages sampling $L_d$ random words from the corresponding vocabulary (we use the same number of words $L_d$ for each document). This should be a very simple problem, since there is neither mixing of words across topics, nor of topics across documents.

Let us compute the log-likelihood, $\log{\mathcal{L}_{\textrm{true}}}$, of the generative model. The process of generating a document works in two step. We first select a language with probability $p(L)$, and we then write a document with probability $p(doc|L)$:

\begin{equation}
\label{likelihood1}
\log{\mathcal{L}_{\textrm{true}}}= \log{p(L)} + \log {p(doc|L) } . 
\end{equation}

Let us focus on the second part,  $ \log {\mathcal{L}^{'}_{\textrm{true}}} = \log {p(doc|L) }$. After we selected the language that we are going to use, every document has the same probability of being generated:

\begin{equation}
\label{likelihood1b}
\log {\mathcal{L}^{'}_{\textrm{true}}} = - L_d \log{N_w} .
\end{equation}

We will also consider $p(L)$ later. We stress that $\log {\mathcal{L}^{'}_{\textrm{true}}}$ is the log-likelihood per document. The symbol $'$ is to recall that the likelihood is computed given that we know which language we are using for the document.

\begin{figure*}
\begin{center}
\includegraphics[width=5in]{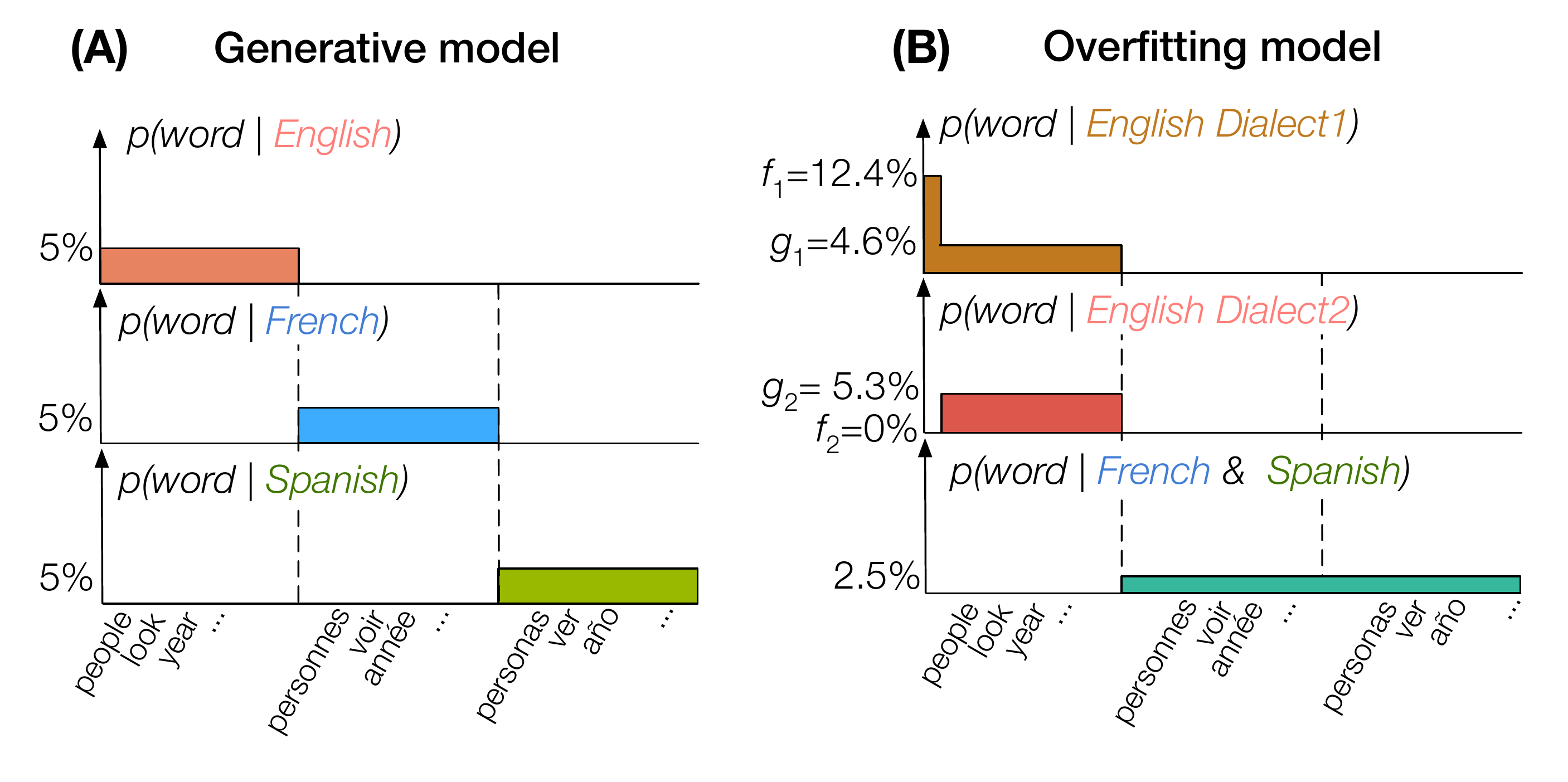}
\caption{Distribution over words for the topics in the generative model (\textbf{A}) and in the most likely model (\textbf{B}). We set $N_w=20$ words in each language's vocabulary. 
}
\label{limit1fig}
\end{center}
\end{figure*}

Now, let us compute the log-likelihood  of an alternative model, where one language (say English) is overfitted in two dialects, and two other languages (say French and Spanish) are merged. Fig. \ref{limit1fig} illustrates how we construct the alternative model. French and Spanish are just one topic, in which each French and Spanish word is equiprobable.
The English words instead are arbitrarily divided in two groups: the first English dialect makes use of words from the former group with probability $f_1$ and words from the second groups with probability $g_1$ and the second dialect has probabilities $f_2$ and $g_2$ for the two groups. We assume that the first group of words is more likely for the first dialect, i.e $f_1 \geqslant g_1$, while the situation is reversed for the second dialect: $g_2 \geqslant f_2$. The general idea is that if a document, just by chance, is using words from the first group with higher probability, it might be fitted better by the first dialect: overfitting the noise improves the likelihood and, if the English portion of the corpus is big enough, this improvement might overcome what we lose by underfitting French and Spanish.

In Sec.~\ref{appendixa}, we prove that the difference between the log-likelihood per English document of the generative model and the alternative model is  bigger than $1/\pi$, regardless of the number of words per document, the size of the vocabulary or the number of documents. More precisely, if $N_w \geqslant L_d$, the difference can also be higher, $ \simeq (\log 2)^2$. Calling ${\mathcal{L}^{'}_E }$, the likelihood per English document in the alternative model, we have that:

\begin{equation}
\label{likelihood2}
\langle \log {\mathcal{L}^{'}_E } \rangle = \log {\mathcal{L}^{'}_{\textrm{true}}}+ C \quad  \textrm{with} \quad C \in [ \sim 0.3, \sim 0.5].
\end{equation}

\begin{figure}
\begin{center}
\includegraphics[width=8cm]{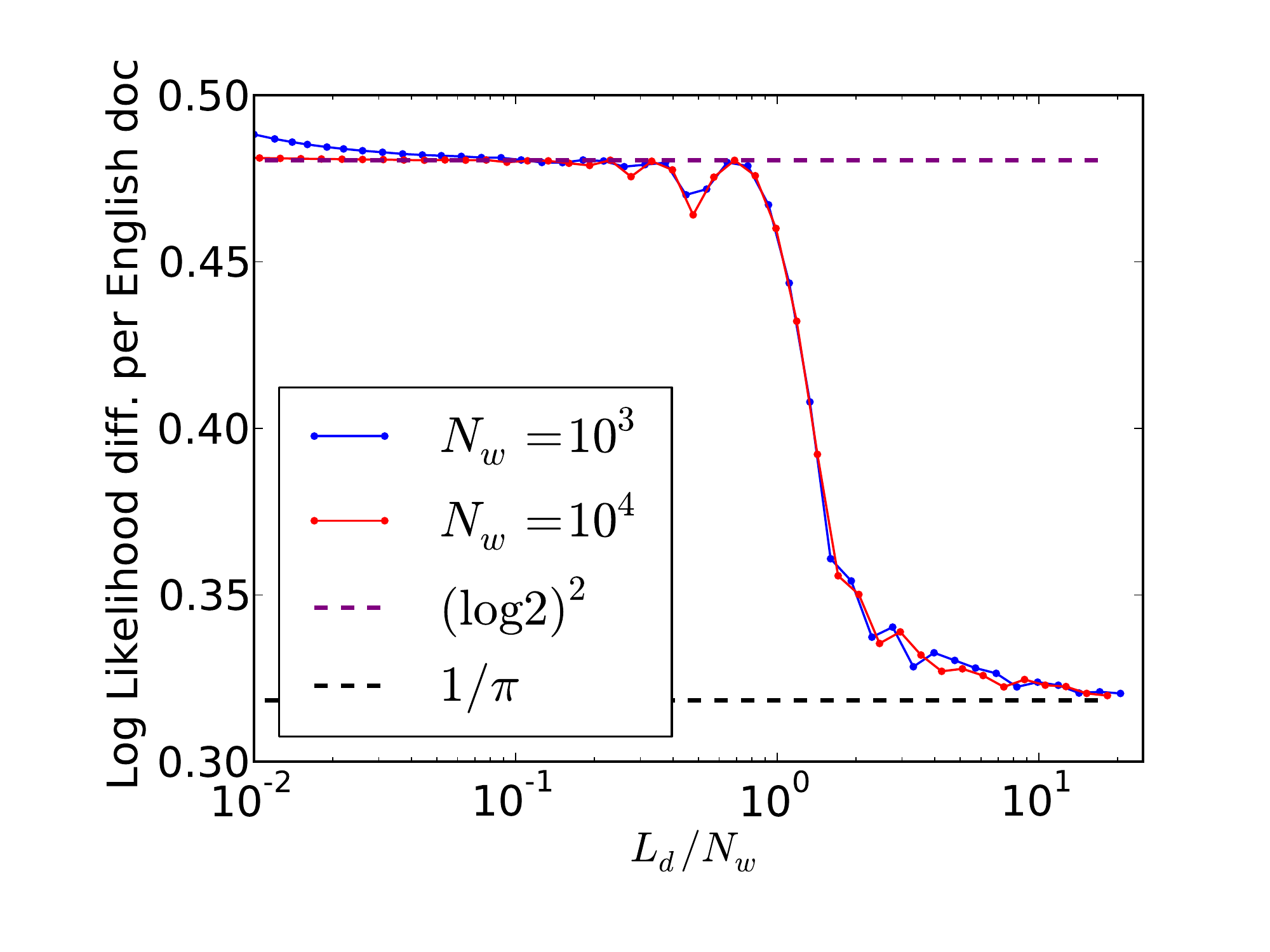}
\caption{Average difference in the log-likelihood per English document of the alternative model and the generative model as a function of the ratio $L_d / N_w$,  (words per document over vocabulary size).  
The function as well as the two dashed lines have been analytically computed in Sec.~\ref{appendixa}. }
\label{limitfig2}
\end{center}
\end{figure}

Fig.~\ref{limitfig2}, shows the log likelihood difference per English document, as a function of $L_d/N_w$.

Keeping the same number of topics, the alternative model  will pay some cost underfitting Spanish and French. Since the languages are merged, the size of the vocabulary is $2 N_w$ and the log-likelihood per Spanish or French document is:

\begin{equation}
\label{likelihood3}
\log {\mathcal{L}^{'}_{SF} } =  \log {\mathcal{L}^{'}_{\textrm{true}}} -L_d \log2.
\end{equation}

Now, to compute the expected log-likelihood of the alternative model we also need to know how often we use the different languages.
Let us call $f_E$ the fraction of English documents, and $f_U$ the fraction of documents written in Spanish or French (underfitted documents).

The average log-likelihood per document of the alternative model can then be written as:

\begin{equation}
\label{likelihood4}
\langle \log {\mathcal{L}^{'}_{\textrm{alt}} } \rangle=   \log {\mathcal{L}^{'}_{\textrm{true}}} +  f_E C - f_U L_d \log2.
\end{equation}

We recall that, so far, we have not considered the probability that each document will pick a certain language, $p(L)$. Symmetric and asymmetric LDA make different assumptions at this point and we treat them both in the next two sections.

\subsection{Symmetric LDA}

PLSA does not account for the probability of picking a language $p(L)$ in the likelihood. LDA instead does consider that: the hyper parameters $\alpha_L$ are a global set of parameters (one per topic) which tune the probabilities that each document is making use of each topic. In our case, each document is uniquely assigned to a language: therefore, for each document, there is a language which has probability 1 and all the other languages have probability 0. This corresponds to the limiting case $\alpha_L = \kappa p(L)$ where the proportionality factor $\kappa$ is very small.

For symmetric LDA, however, all the $\alpha_L$ are equal. This implies that, regardless of the actual size of the languages, the algorithm fits the data with a model for which $p(L)=1/K$ (we recall that $K$ is the number of languages). Therefore:

\begin{equation}
\label{final1}
\log{\mathcal{L}_{\textrm{true}}}= - \log{K} -L_d \log N_w \quad \textrm{and} 
\end{equation}
\begin{equation*}
\quad \langle \log {\mathcal{L}_{\textrm{alt}} } \rangle=   \log {\mathcal{L}_{\textrm{true}}} +  f_E C - f_U L_d \log2.
\end{equation*}

 If $f_E$ is big enough, the likelihood of the alternative model can be higher than the one of the generative model.
To be more concrete, let us consider an example. If $L_d=10$ and $N_w=20$, in Sec.~\ref{appendixa}, we show that $C$ can be as high as  $ \simeq 0.476$. 
Let us consider the simplest case of just three topics, $f_U=1-f_E$. Setting the right hand side of Eq.~\ref{final1} to zero, we find that if $f_E \geqslant 0.936$ the alternative model has a better likelihood. If the topics are not balanced enough, symmetric LDA cannot find the right generative model, regardless of the absence of any sort of mixing. However, this critical value actually depends on $L_d$, and increasing $L_d$ the generative model will eventually get a better likelihood. The case $L_d \gg 1$, is treated in detail below.

\subsection{Asymmetric LDA}

For asymmetric LDA, the average log-likelihood of the true model becomes:

\begin{equation}
\label{likelihood_full}
\langle \log {\mathcal{L}_{\textrm{true}}} \rangle = -H_{\textrm{true}} + \log {\mathcal{L}^{'}_{\textrm{true}}} ,
\end{equation}

where $H_{\textrm{true}}$ is the entropy of the language probability distribution, $H_{\textrm{true}}= - \sum_L p(L) \log{p(L)}$.

For the sake of simplicity, let us assume that French and Spanish are equiprobable, as well as the two English dialects (see Sec.~\ref{appendixa}). For the alternative model:

\begin{equation}
H_{\textrm{alt}}  =  H_{\textrm{true}} + (f_E  - f_U)\log{2} .
\end{equation}

From Eq.~\ref{likelihood4}, we finally get:

\begin{equation}
\label{final2}
\langle \log {\mathcal{L}_{\textrm{alt}} } \rangle=  \langle \log {\mathcal{L}_{\textrm{true}}} \rangle -  f_E (\log2-C) - f_U (L_d-1) \log2 .
\end{equation}

Since $\log2>C$, now the generative model actually has the highest likelihood: in principle, asymmetric LDA is always able to find the generative model. 
The ratio of the two log-likelihoods, if the documents are long enough, becomes:

\begin{equation}
\label{final_ration}
\frac{\langle \log {\mathcal{L}_{\textrm{alt}} } \rangle} { \langle \log {\mathcal{L}_{\textrm{true}}} \rangle} = 1  -  \frac{ f_U  \log2 }{\log N_w}.
\end{equation}

The same equation holds for symmetric and asymmetric LDA, as well as PLSA. Therefore, even if we had infinite amount of information (infinite number of documents and words per document), the ratio of the two likelihoods can actually be very close to 1.

\subsection{Finding the generative model in practice}

The number of alternative models is huge. In Sec.~\ref{appendixa}, we show that if each language had a vocabulary size $N_w=1000$,
we can find $\sim K \times 10^{300}$ alternative models (this is a conservative estimate): assuming $f_U=0.2$ (which would correspond to $10$ equiprobable topics), the relative difference in their log-likelihood is $\sim 2\%$ as we can estimate from Eq. \ref{final_ration}.

\begin{figure*}
\begin{center}
\includegraphics[width=5in]{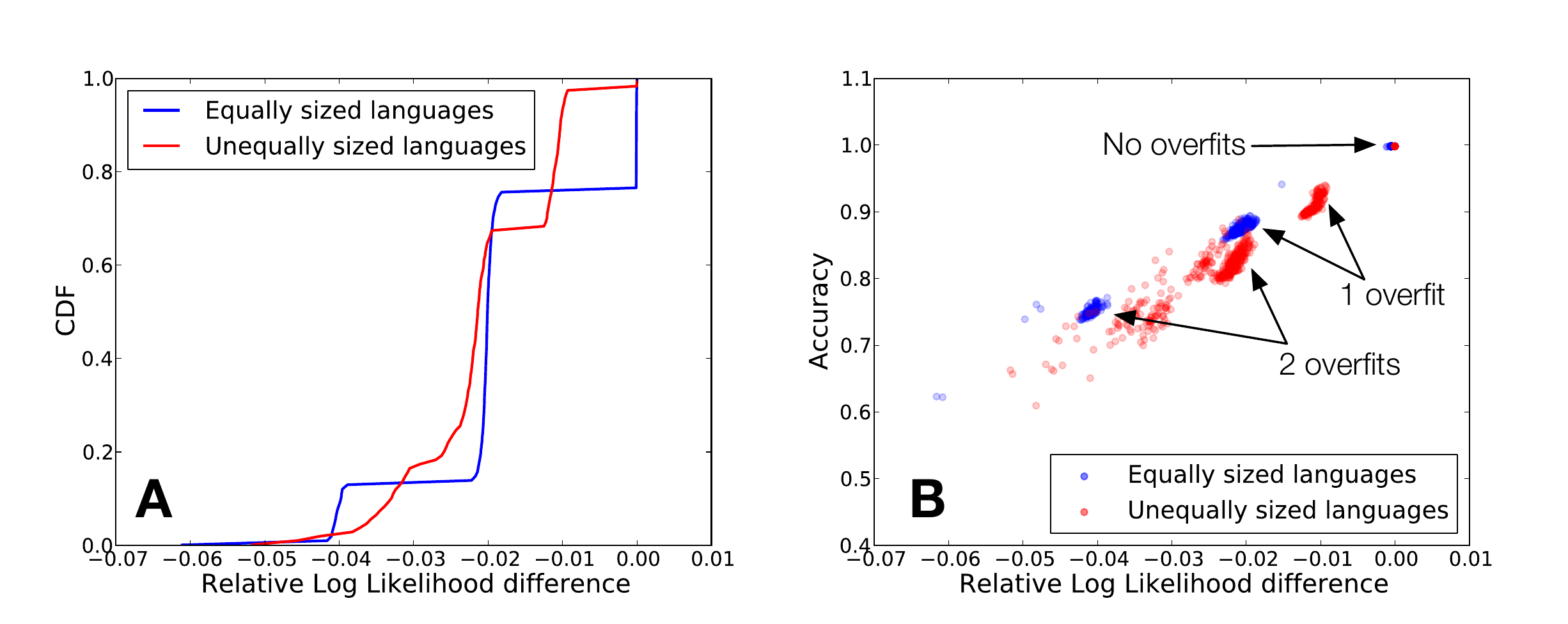}
\caption{In this test, the corpus has 5000 documents of 100 words each, and the vocabulary of each language has 1000 equiprobable words. In the equally sized case, we consider 10 equiprobable languages, while in the heterogeneous case, we considered $2$ languages with probability $30\%$ each, and 8 languages with probability $5 \%$. \textbf{A.} Cumulative probability of  the relative difference of the log likelihood of the generative model and the one found by the algorithm. \textbf{B.} Scatter plot of the relative difference of the log likelihood versus the accuracy of the algorithm (accuracy is the Best Match similarity of the two models, see main text). Clear clusters are visible according to the how many languages are overfitted.
Fig.~2 of the main paper, supports the same conclusion also after we removed the assumption that words are equiprobable.
}
\label{basin}
\end{center}
\end{figure*}

One might argue that, even if the relative difference of the log-likelihood is small, we have not considered that the basin of attraction of the generative model can be very large, so that optimization algorithms might actually be very effective in finding it anyway. Fig.~\ref{basin} shows that the probability of finding the correct model for equiprobable languages is $\sim 20\%$, while in the heterogeneous case is $\sim 2\%$ (this was computed using variational inference  \cite{blei2003LDA}).

\subsection{Model competition in hierarchical data}
\label{seb_sec::modelcomp}

In the previous sections, we only discussed the difference in likelihood  of the generative model and an alternative model with the same number of topics $K$. In this section, we consider a similar test case for which, however, we fit the data with a model with $K-1$ topics.

\begin{figure}
\begin{center}
\includegraphics[width=3in]{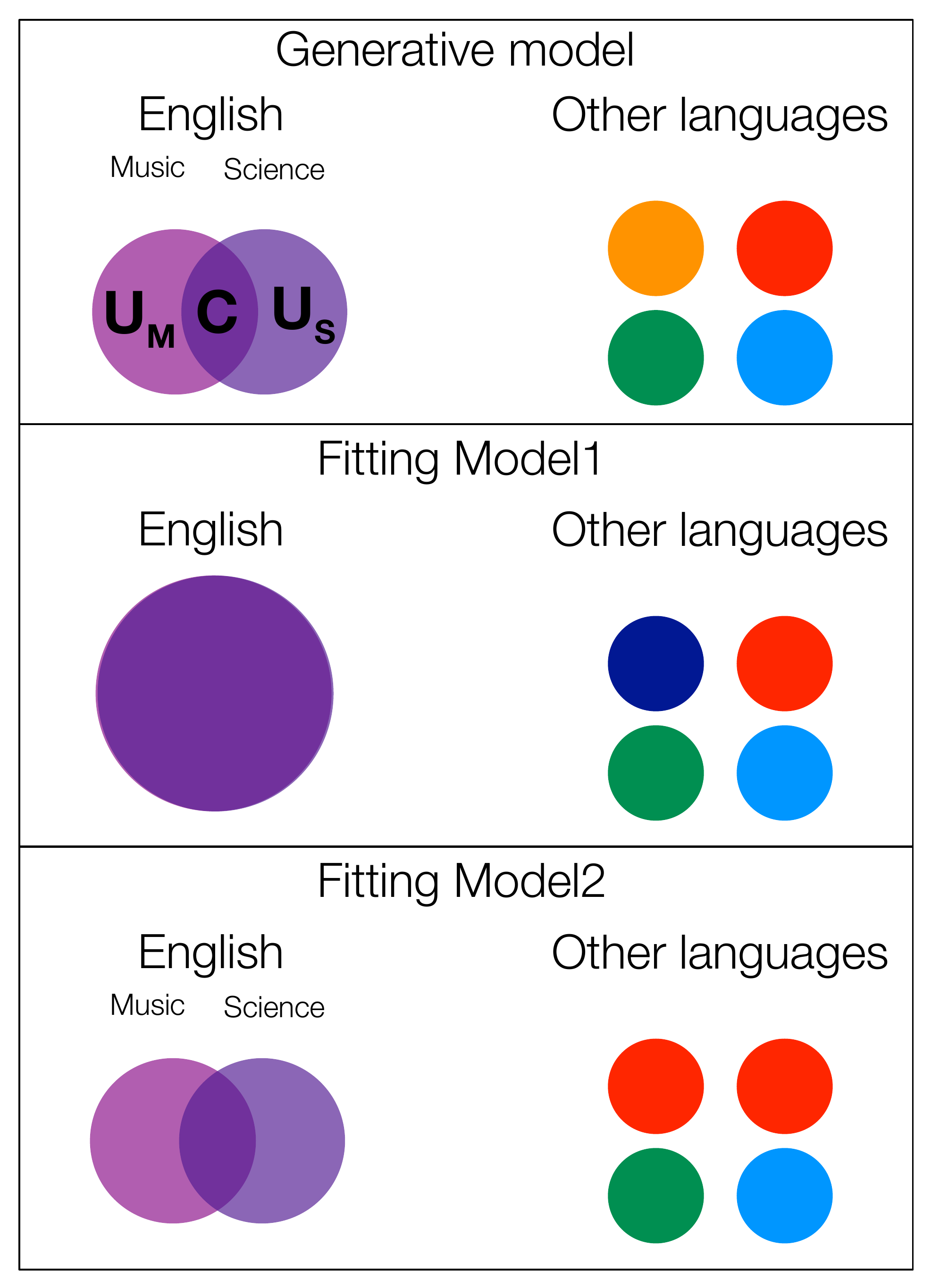}
\caption{Generative model and two compiting models. In this example, we have $K-1$ languages but one language (English) is bigger than the others and have two subtopics (``music" and ``science"). $U_M$ is the number of words in the English vocabulary which can only be found in the music subtopics, $U_S$ is the equivalent for science, whereas $C$ is the number of common words between the two subtopics.  If many documents are written in English, Model 2 has a better likelihood than Model 1.}
\label{kminusone}
\end{center}
\end{figure}

The generative model we consider here is illustrated in Fig~\ref{kminusone}: we have $K-1$ topics which have no words in common with any other topic and one bigger topic, say English, which has two subtopics, say ``music'' and ``science'', which share some words. Let us call $U_M$ the number of words in one of the English subtopics (music)  which cannot be found in the other subtopic, $U_S$ the number of words which can only be found in the other subtopic (science), and $C$ the number of words in common between the two subtopics.  We further assume that $U_M=U_S=U$, the subtopics are equiprobable, and given a subtopic, each word is equiprobable. Let us call $N_w$ the number of words in each non-English language, $p_E$ the fraction of English documents and $p_k$ the fraction of documents written in a different language (for sake of simplicity, all languages but English are equiprobable). 

This model should be fitted with $K$ topics. However, let us assume that we do not know the exact number of topics (as it is usually the case) and we try to fit the data with $K-1$ topics. In Fig.~\ref{kminusone} we show two possible competing models: the first model correctly finds all the languages, while the second correctly finds the English subtopics but merges two languages. 

With similar calculations as above, we can prove (see Sec.~\ref{deghier}) that the first model has higher likelihood if:

\begin{equation}
\label{modelcomeq}
2 p_k > p_E \frac{U}{C+U}.
\end{equation}

The previous equations holds for symmetric LDA, and also asymmetric LDA if $L_d \gg 1$ (the exact expression for  asymmetric LDA can be found in Sec.~\ref{deghier}).  If $U=0$, the first model is always better (there are no subtopics), if $C=0$, one model is better than the other if it under-fits the smaller fraction of documents. In general, if English is used enough and $U>0$, the second model better fits the data.

Let us consider a numerical example: consider $p_E=50\%$, $U_M=U_S=50$ words and $C=900$ words ($1,000$ total words in the English vocabulary). This means that $90\%$ of the English words are used by both subtopics. Eq.~\ref{modelcomeq} tells us that we are going to split English in the two subtopics, if there are two other topics to merge with $2p_k< 2.6\%$.

We believe that this is the basic reason why big journals such as Cell and Astronomical journals are split by standard LDA in the Web of Science dataset (see Sec.~\ref{likelihood::sec}). In general, since real-world topics are likely to display a hierarchical structure similar to the one described here, we argue
that heterogeneity in the topic distribution makes standard algorithms prone to find subtopics of large topics before resolving smaller ones.

\section{A network approach to topic modeling}
\label{network_sec}

We give here a detailed description of TopicMapping.
The method works in three steps.

First, we build a network of words, where links connect terms appearing in the same documents more often than what we could expect by chance. Second, we define the topics as clusters of words in such a network, using the Infomap method \cite{rosvall2008maps} and then we compute the probabilities $p(topic|doc)$ and $p(word|topic)$ locally maximizing a PLSA-like likelihood. Finally, we can refine the topics further optimizing the (asymmetric) LDA likelihood via variational inference \cite{blei2003LDA}.

\paragraph{How to define the network.}

A corpus can be seen as a weighted bipartite network of words and documents: every word $a$ is connected to all documents where the word appears. The weight $\omega_a^d$ of the link  is the number of times the word is repeated in document $d$.

From this network, we would like to define a unipartite network of words which have many documents in common. A very simple measure of similarity between any pair of words $a$ and $b$ is the dot product similarity:

\begin{equation}
\label{dotproduct_similarity}
z_{a,b} = \sum_d \omega_a^d \times \omega_b^d .
\end{equation}

From this definition, it is clear that generic words, like ``to" or ``of", will be strongly connected to lots of more specific words, putting close terms related to otherwise far semantic areas. A possible way to filter out generic words is to compare the corpus to a simple null model where all words are randomly shuffled among documents.

For this purpose, we need to consider the probability distribution $p(z_{a,b})$ of the dot product similarity defined in Eq. \ref{dotproduct_similarity}.
We start considering that in the null model each weight $\omega_a^d$ is now a random variable which follows a hypergeometric distribution with parameters given by: the total number of words in document $d$, $L_d$,  the total number of occurrences of word $a$ in the whole corpus, $s_a = \sum_d \omega_a^d$, and the total number of words in the corpus $L_C= \sum_d L_d$. The mean $\langle \omega_a^d \rangle$ is:

\begin{equation}
\label{mean_weight_eq}
\langle \omega_a^d \rangle = \frac{L_d \times s_a} {L_C} .
\end{equation}

Assuming a large enough number of documents,  we can neglect the correlations among the variables $\omega_a^d$ and, from Eqs.  \ref{dotproduct_similarity} and  \ref{mean_weight_eq}, we get:

\begin{equation}
\label{poissonian_approx}
\langle z_{a,b} \rangle =  \sum_d \langle \omega_a^d \rangle \times \langle \omega_b^d \rangle =  \frac{  s_a s_b} {L_C^2} \sum_d L_d^2.
\end{equation}

\begin{figure}[h!]
\begin{center}
\includegraphics[width=8cm]{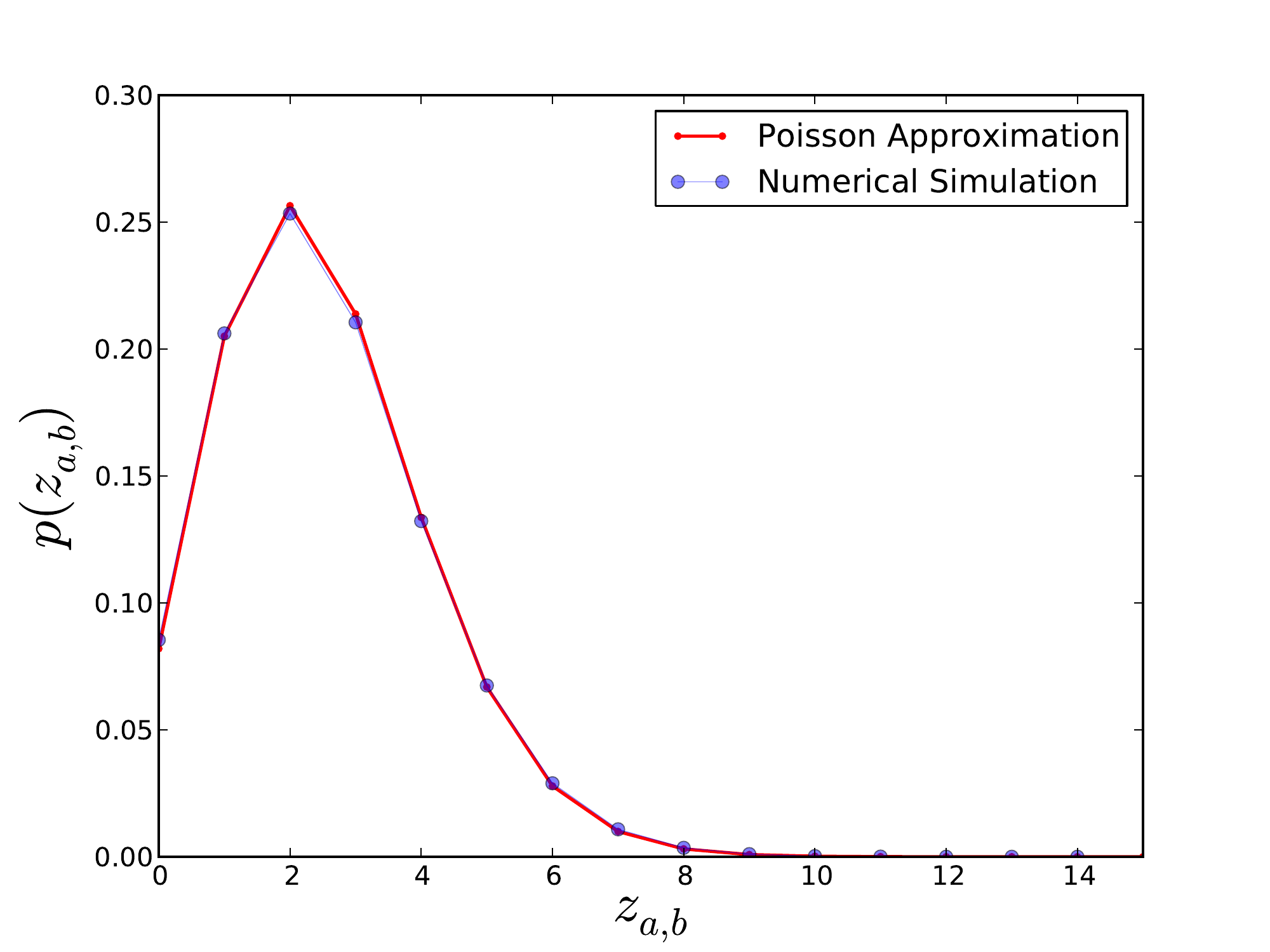}
\caption{Poisson approximation of  the probability distribution $p(z_{a,b})$ of the dot product similarity of words $a$ and $b$ in a randomly shuffled corpus. The occurrences of the words are $s_a=10$, $s_b=200$, and there are $1000$ documents of length drawn uniformly between $10$ and $100$ words.}
\label{SI_fig2}
\end{center}
\end{figure}

Since $z_{a,b}$ is the sum of rare events (if $L_C \gg 1$), its probability distribution can be well approximated by a Poisson distribution $\textrm{Pois}_{\langle z_{a,b} \rangle}(z)$ with average given by Eq. \ref{poissonian_approx}, as shown in Fig. \ref{SI_fig2}.

Finally, our procedure to filter out the noise consists in fixing a $p$-value, and for all pairs of words $a$ and $b$ which share at least one document, we compute $z_{a,b} - Z_p(s_a, s_b)$, where the latter term 
is the $(1-p)$-quantile of the Poisson distribution $\textrm{Pois}_{\langle z_{a,b} \rangle}(z)$. Being more precise, $ Z_p(s_a, s_b)$ 
is the largest non significant dot product similarity:

\begin{equation}
\label{poissonian_quantile}
Z_p(s_a, s_b) = \max_x \big \{x \textrm{ such that: }  \sum_{z=x}^\infty \textrm{Pois}_{\langle z_{a,b} \rangle}(z) > p \big \} .
\end{equation}

\begin{figure*}[ht!]
\begin{center}
\includegraphics[width=5in]{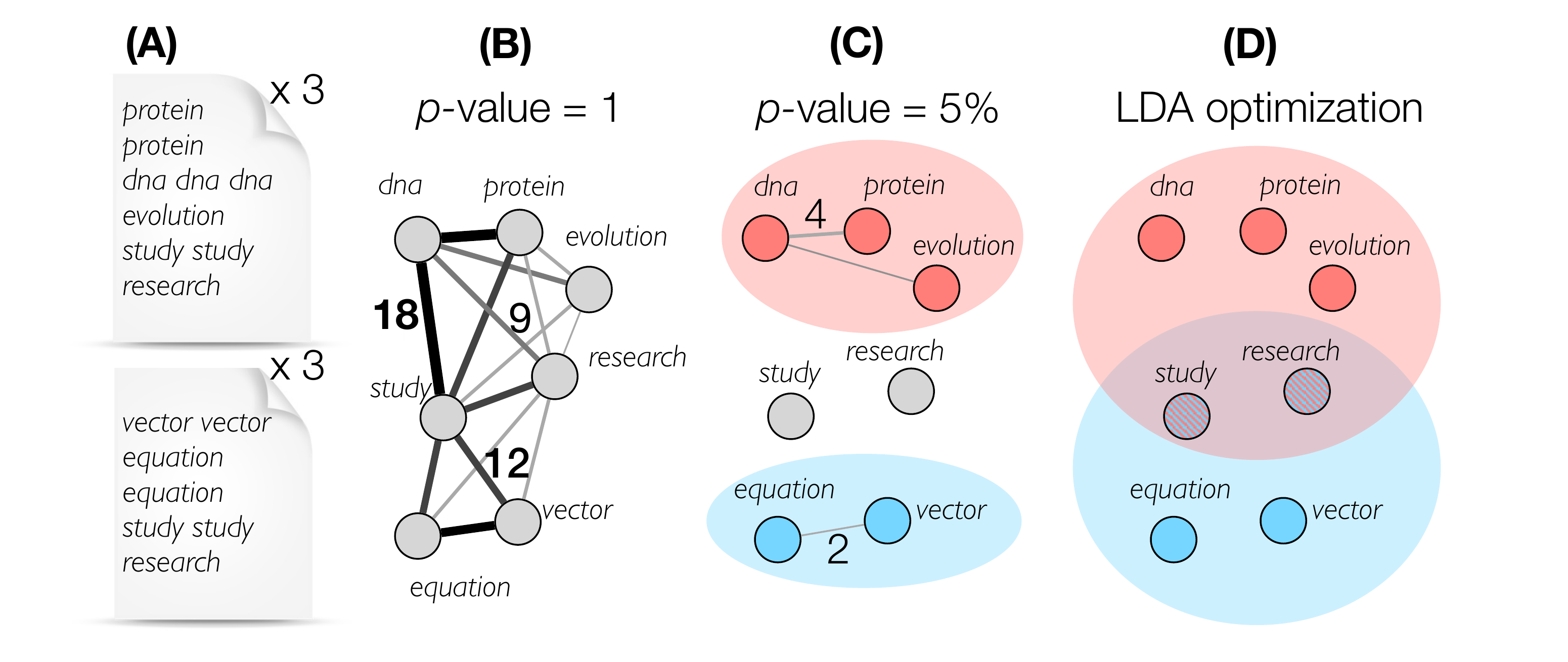}
\caption{{\textbf{A}. The corpus comprises six documents, 3 are about biology and 3 about math.   
\textbf{B}. We build a network connecting words with weights equal to their dot product similarity. 
\textbf{C}. We filter non-significant weights, using a $p$-value of $\%5$. Running Infomap \cite{rosvall2008maps}
on this network, we get two clusters and two isolated words (\textit{study} and \textit{research}).
\textbf{D.} We refine the word clusters using a topic model: the two isolated words can now be found in both topics.
}
\label{fig::map_ill}
}

\label{figmap}
\end{center}
\end{figure*}

$z_{a,b} - Z_p(s_a, s_b)$ is the weight of the link between words $a$ and $b$, if positive.
Fig.~\ref{fig::map_ill} shows an example.

\paragraph{Finding the topics as clusters of words and Local Likelihood Optimization.}

Once the network is built, we detect clusters of highly connected nodes using the Infomap method \cite{rosvall2008maps}. This provides us with a hard partition of words, meaning that words can only belong to a single cluster.

We now discuss how we can compute the distributions $p(topic|doc)$ and $p(word|topic)$, given a partition of words.

We recall that in the probabilistic model of how documents are generated, we assume that every word $w$ appearing in document $d$ has been drawn from a certain topic. We are in the realm of the \textit{bag of words} approximation, and therefore we are completely discarding any information about the structure of the documents. Then, it is reasonable to assume that every time we see a certain word in the same document, it was always generated by the same topic: let us denote this topic as $\tau(w, d)$.

We identify the topic $\tau(w, d)$ with the single module where word $w$ is located by Infomap, $\tau(w)$: in fact, since the partition is hard (no words can sit in different modules), there is no dependency on the documents. Therefore, $p(t|w)=\delta_{t,\tau(w)}$ and:

\begin{equation}
\label{pmodel}
p(w,t)=p(w) \, \delta_{t,\tau(w)}   \,\,\,\,\,\, \textrm{and} \,\,\,\,\,\, p(t|d) = \frac{1}{L_d} \sum_w \omega_w^d  \, \delta_{t,\tau(w)} .
\end{equation}

It is also useful to introduce $n(w,t) = L_C \, p(w,t)$, which is the number of times topic $t$ was chosen and word $w$ was drawn. 

So far, we have got a model where all words are very specific to topics and documents use many topics, which is probably far from being a good candidate generative model.
The model can be substantially improved optimizing the PLSA-like likelihood:

\begin{equation}
\label{likelihood_plsa_model_equation}
\mathcal{L} = \prod_{w,d} p(w,d) = \prod_{w,d} \sum_t  p(w|t) \, p(t|d) \, p(d) .
\end{equation}

We then describe a series of very local moves aimed at improving the likelihood of the model. 
The local optimization algorithm aims at fuzzing the topics and making documents more specific to fewer topics. For that, it simply finds, for each document,  topics which are infrequent (more precise definition follows) and ``move"  the words drawn from that topic to the most important one in that document. 

\begin{enumerate}

  \item For each document $d$, we find its most significant topic, $\tau_d$: this is done selecting the topic with the smallest $p$-value, considering a null model where each word is independently sampled from topic $t$ with probability $p(t)=\sum_w p(w) p(t|w)$. Calling $x$ the number of words which actually come from topic $t$, ($x= L_d \times p(t|d)$, see Eq.~\ref{pmodel}),  the $p$-value of topic $t$ is then computed using a binomial distribution, $\textrm{B}(x;  L_d, p(t))$.

  \item For document $d$, we define the \textit{infrequent} topics $t_{in}$ as those which are used with probability smaller than a parameter:   $p(t_{in}|d)<\eta$.  
 
   We consider the most significant topic $\tau_d$ (see above) and we increment $p(\tau_d|d)$ by the sum of the probabilities of the infrequent topics, while all $p(t_{in}|d)$ are set to zero. Similarly, $n(w,t)$ has to be decreased by $\omega_w^d$ for each word $w$ which belongs to an infrequent topic, and $n(w,\tau_d)$ is increased accordingly.
  \item We repeat the previous step for all documents. We then compute $p(w,t) = n(w,t) / L_C$, as well as the the likelihood of the model, $\mathcal{L}_\eta$, where we made explicit its dependency on $\eta$.
  \item We loop over all possible values of $\eta$ (from $0\%$ to $50\%$ with steps of $1\%$) and we pick the model which maximizes  $\mathcal{L}_\eta$.

\end{enumerate}

\paragraph{LDA Likelihood optimization.}

The model we find, at this point, can be refined further via
iterations of the Expectation-Maximization algorithm optimizing the LDA likelihood.
The algorithm follows closely the implementation from \cite{blei2003LDA}.
The main difference, however, is that, for computing efficiency, we use sparse data structure,
where words and documents are assigned to only a subset of the topics.

In most cases, the model does not change very much and the algorithm converges very quickly. However, if topics are very heterogenous in size, we might encounter situations similar to the one described in Sec.~\ref{seb_sec::modelcomp} (see Sec.~\ref{wikipedia} for an example). In practice, the software records models every few iterations, allowing users to better explore the data.

\paragraph{Implementation details.} 

Here, we would like to make a few points more precise.

\begin{enumerate}
	
	\item The filtering procedure and the LDA likelihood optimization in TopicMapping are deterministic. 
	Instead, optimizing Infomap's code length uses a Monte Carlo technique, which can be performed multiple times. 
	The number of runs for  Infomap's optimization was set  to 10 in most tests, although most results barely change with a single run.
	For measuring the reproducibility in Sec.~\ref{wos_hierarchy}, instead, we used 100 runs, because the topic structure
	 is less sharp and we need some more runs to achieve good reproducibility (each run takes about a minute).
	\item After running Infomap, we might find that some words have not been assigned to any topics, because all their possible connections to other words have not been considered significant. In each document which uses any of them, we automatically assign these words to its most significant topic, $\tau_d$. 
	\item Some (small) topics might have not been selected as the most significant by any document.
	We remove these topics before the filtering procedure: if we do not, high values of the filter $\eta$ will 
	yield models where these topics do not appear at all, and this might penalize their likelihood just because the number of topics is diminished.
	\item Depending on the application, it might also be useful to remove very small topics
	even if they were selected as the most significant by a handful of documents
	(this is especially important to avoid the following LDA optimization to inflate them, see Sec.~\ref{lda_initial_conditions}).
	We used no threshold for the synthetic datasets, but we selected a threshold of 10 documents for the journals in Web of Science,
	and 100 documents for Wikipedia.
	In the implementation of the software, we let the users choose a threshold for removing small topics.
	\item The initial $\alpha$ for LDA optimization was set to $0.01$ for all topics.
	
\end{enumerate}

\section{Held-out likelihood and effective number of topics}
\label{heldout_likelihood}

\begin{figure}
\begin{center}
\includegraphics[width=3.1in]{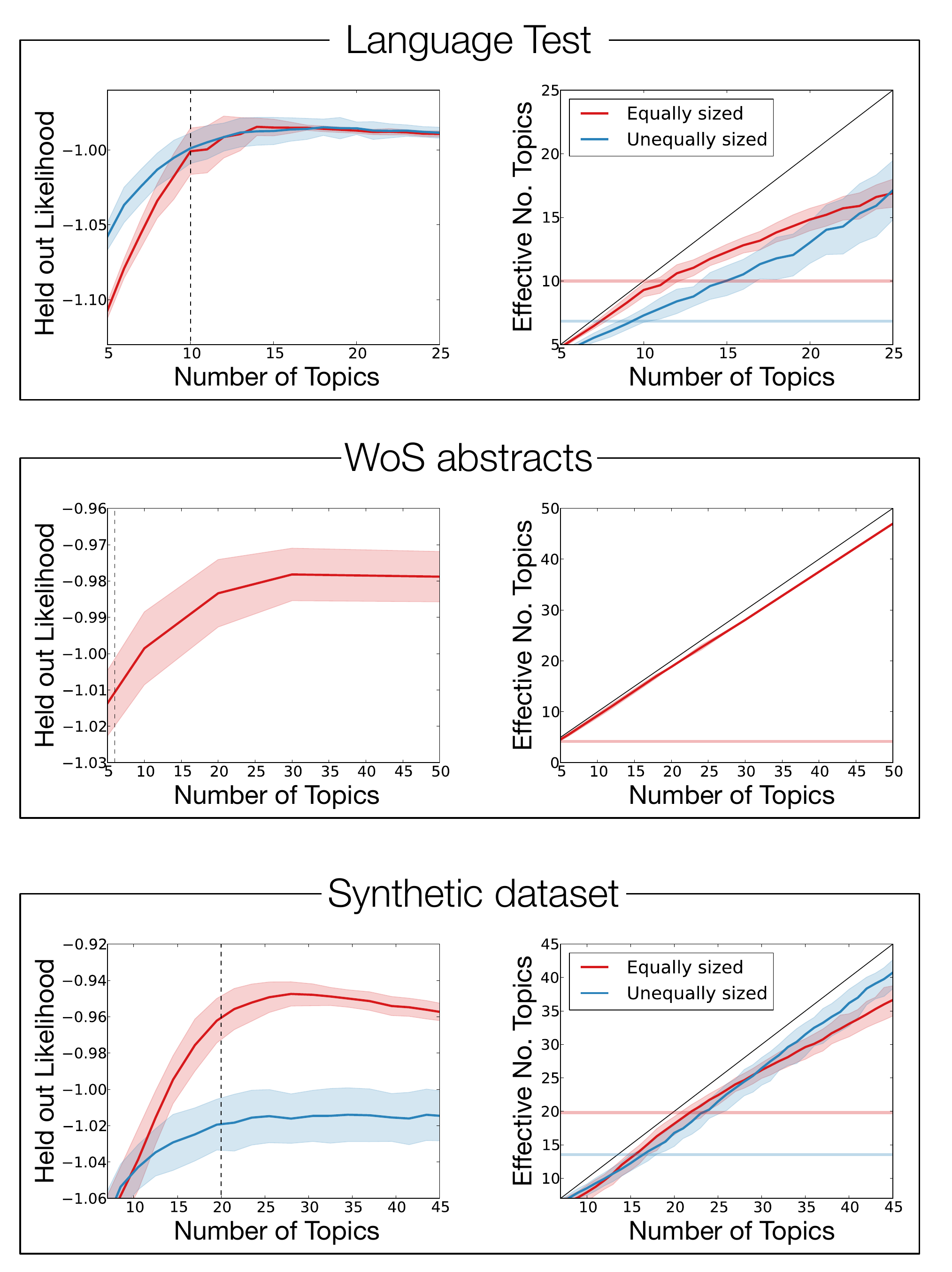}
\caption{Held-out likelihood and effective number of topics for the three datasets we considered in the main paper. In the language test, we considered $5,000$ documents, while, in the synthetic dataset, we set  $\alpha=10^{-3}$ and  the fraction of generic words to $25 \%$. The dashed black lines on the left indicated the number of topics $K$ that should have been selected by the method. The black line on the right-hand panels is $y=x$ (the highest achievable value of the effective number of topics) and the horizontal lines are the actual effective number of topics. }
\label{neff_fig}
\end{center}
\end{figure}

The most used method for selecting the right number of topics, consists in ($i$) holding out a certain fraction of documents (say 10\% of the corpus), ($ii$) training the algorithm on the remainder of the dataset,  ($iii$) measuring the likelihood of the held-out corpus for the model obtained on the training set. The best number of topics should be the one for which the held-out likelihood is maximum.
Fig.~\ref{neff_fig} shows that  this method tends to give a higher number of topics that the actual one.

We also show that LDA tends to provide models in which $p(topic)$ is fairly close to a uniform distribution.  To assess this, we compare the entropy of the topic distribution,

\begin{equation}
h(p_t) = - \sum_{topic=1}^K p(topic) \log_2 \,p(topic) , 
\end{equation}

with the maximum possible entropy, i.e. those achieved by equally probable topics: $h_u= \log_2(K)$. In fact, it is easier to compare the exponential entropy \cite{campbell1966exponential} of the topic probability distributions: $2^{h(p_t)}$ versus $K$. The former can be seen as an effective number of topics: it is the number of topics needed by a uniform distribution to achieve the same entropy. Fig.~\ref{neff_fig}  shows that indeed, the effective number of topics is rather close to the input $K$.

\section{Additional analysis on the synthetic datasets}
\label{additional_sec}

In this section, we present five supplementary sets of results related to the synthetic datasets, presented in Fig.~4 in the main paper. In the first section, we measure the performance of the algorithms in terms of perplexity \cite{blei2003LDA} (a standard measure of quality for topic models) and we show that, for our case, this evaluating method has a fairly low discriminatory power. We then propose a visualization of the comparison between the correct generative model and the ones found by the algorithms we considered. The third section is dedicated to measuring the performance of the methods in case we do not have information about the correct number of topics to input. In the fourth section, we study how the performance of LDA is affected by the initial conditions of the optimization procedure, and we show that they are crucial, as expected. Finally, we compare the performance of TopicMapping before and after running LDA as a refinement step.

\subsection{Perplexity}

\begin{figure}[h!]
\begin{center}
\includegraphics[width=8cm]{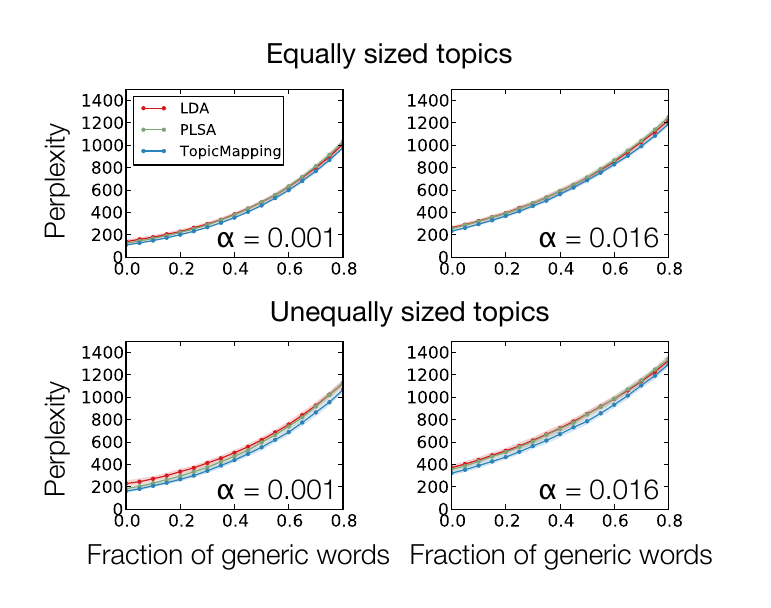}
\caption{
Evaluating the performance of several algorithms on synthetic corpuses measuring perplexity for several values of the parameters (the other parameters are the same as in Fig.~4 in the main paper). Perplexity seems to have low discriminatory power in this test.
}
\label{fig3perple}
\end{center}
\end{figure}

Fig.~\ref{fig3perple} shows the performance of the algorithms on the synthetic datasets in terms of perplexity (in Sec.~\ref{appendix_perple} we explain in detail how perplexity is defined). Algorithms which yield a lower perplexity are considered to achieve a better performance because the model they provide is less ``surprised" by  a portion of the datasets which they have never seen before. The advantage of this approach is that it can be implemented for generic real-world datasets, where the actual generative model is unknown. However, in the study of our interest, the measure performs poorly in discriminating the methods.

\subsection{Visualizing topic models}

\begin{figure}[h!]
\begin{center}
\includegraphics[width=8cm]{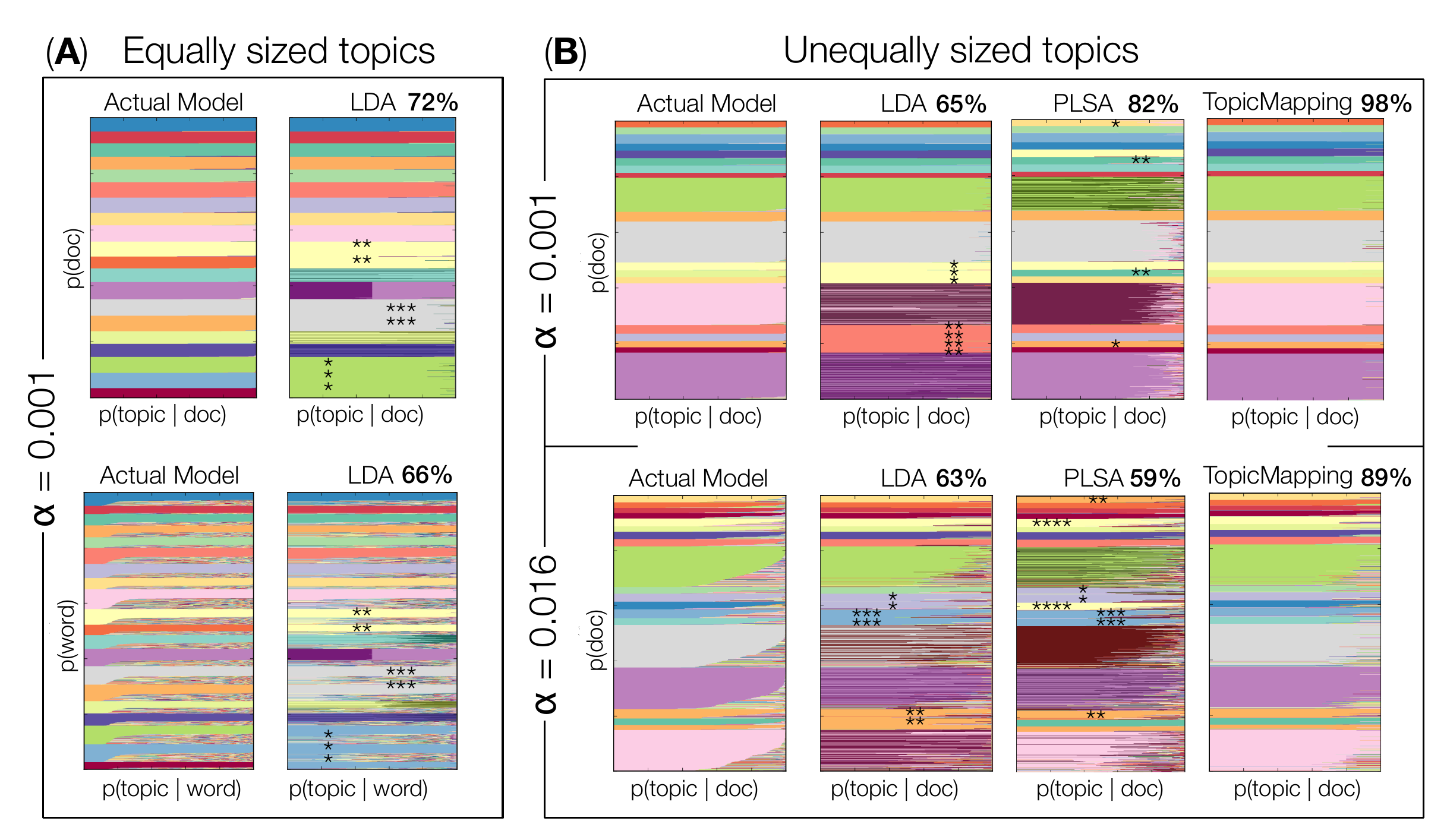}
\caption{
Topic comparison for the synthetic datasets. All parameters are the same as in Fig.~4 in the main paper, and we set the fraction of \textit{generic} words (words which are used uniformly across documents) to $40\%$.
Every rectangle is split in $1000$ horizontal bars, one for document. Each bar is divided in color blocks representing topics, with block size proportional to $p(topic|doc)$. The documents are sorted according to their most prominent topic. 
\textbf{A.}  Performance of LDA, for equally sized topics and  $\alpha=0.001$. 
The ``*" symbols indicate topics inferred by LDA in which two or more actual topics are merged. Top: comparison for documents. Bottom: same procedure for words: generic words are clearly distinguishable from specific ones. The numbers on the corners are obtained from the topic similarity (see main text).
\textbf{B.}  Unequally sized topics. We show results for two values of $\alpha$, $0.001$ and $0.016$. Comparison of documents only is shown. We compare LDA, PLSA and TopicMapping.
}
\label{figviz}
\end{center}
\end{figure}

Fig.~\ref{figviz} shows a visualization of the performance of the methods on the synthetic datasets. We selected a few runs where the algorithms have got an average performance. The colors allow to show in which way standard LDA and PLSA fail in getting the generative model. Similarly to what happens in the language test, some (small) topics are merged together (indicated by a ``*" symbol) and some other topics are overfitted in two or more dialects.

\subsection{Performances for different number of topics.}

Here we discuss how the performance of LDA and PLSA changes if we do not know the exact number of topics. In the main paper, we have fed the algorithms the right number of topics, although we have shown (Sec.~\ref{heldout_likelihood}) that it is hard to guess this information. Here, we show what we get setting a different number of topics, but still reasonably close to the right value ($ K= 20$). In general, the performance gets worse as we move further from the correct number, although 15 or 25 topics sometimes give slightly better results.
We also show that the results do not change very much if we increase the number of documents to $5,000$.

\begin{figure}[h!]
\begin{center}
\includegraphics[width=8cm]{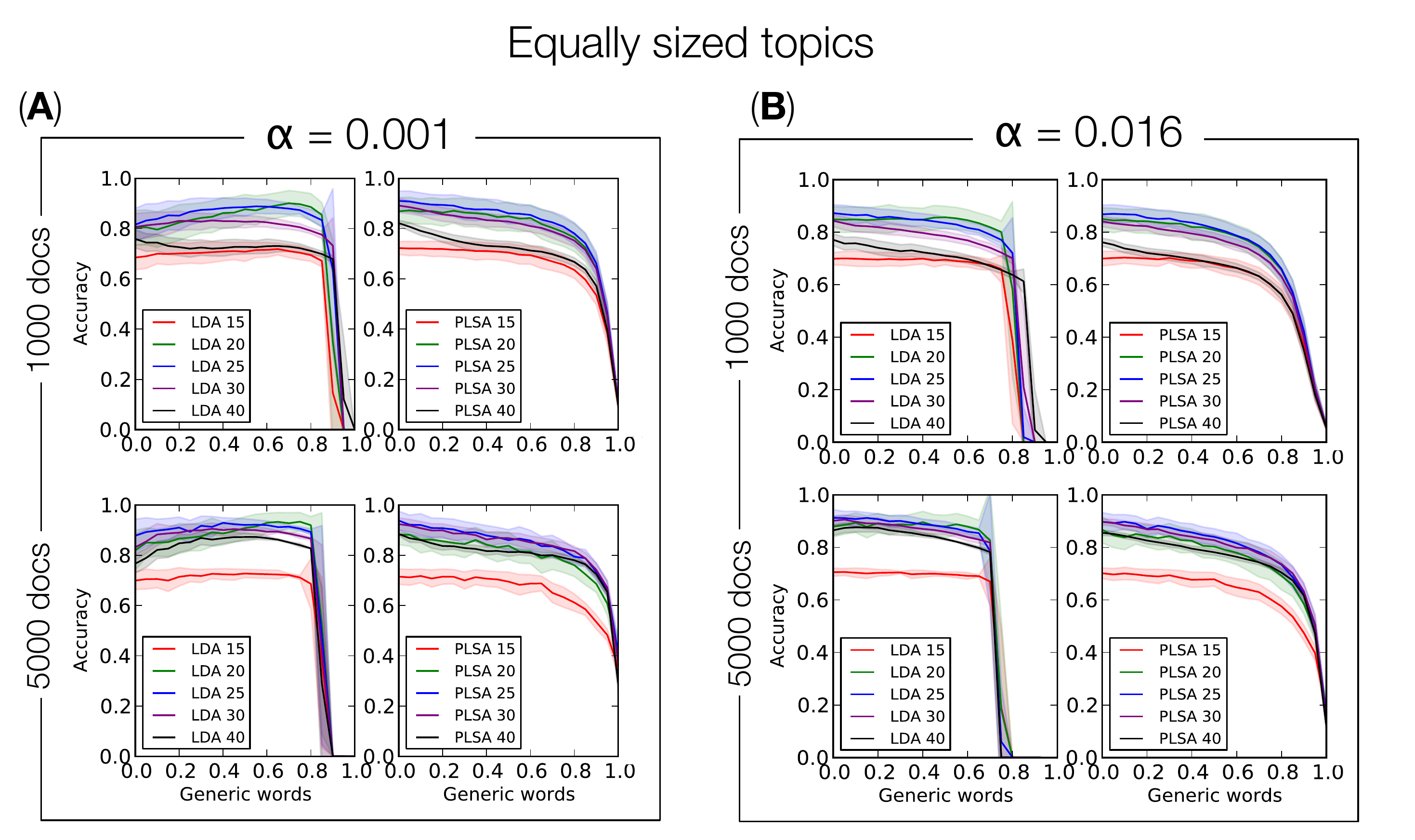}
\includegraphics[width=8cm]{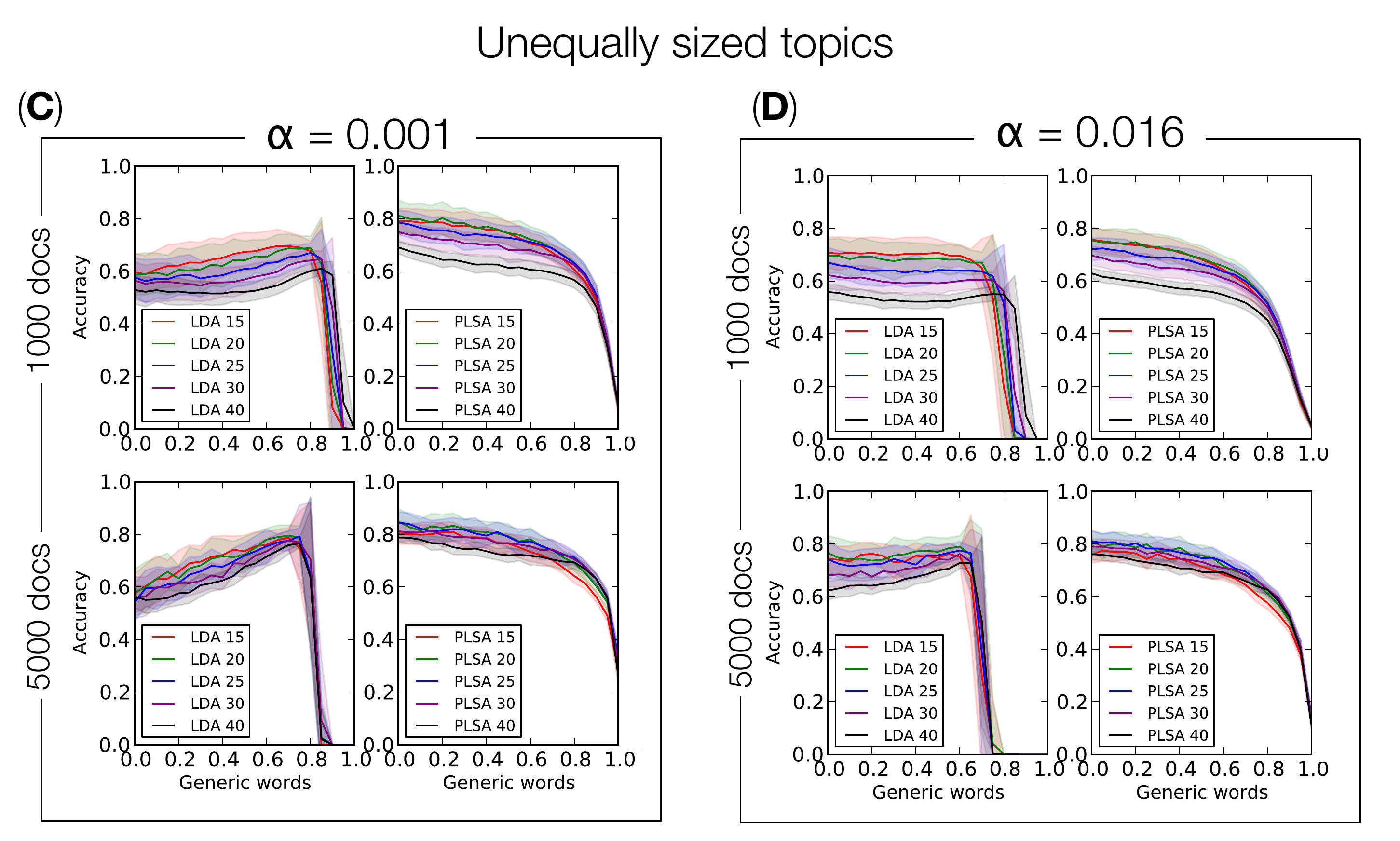}
\caption{Performance of LDA and PLSA when we input different number of topics. The number of topics in the generative model is 20.}
\label{10times_fig}
\end{center}
\end{figure}

\subsection{LDA initial conditions}
\label{lda_initial_conditions}

\begin{figure}[h!]
\begin{center}
\includegraphics[width=6cm]{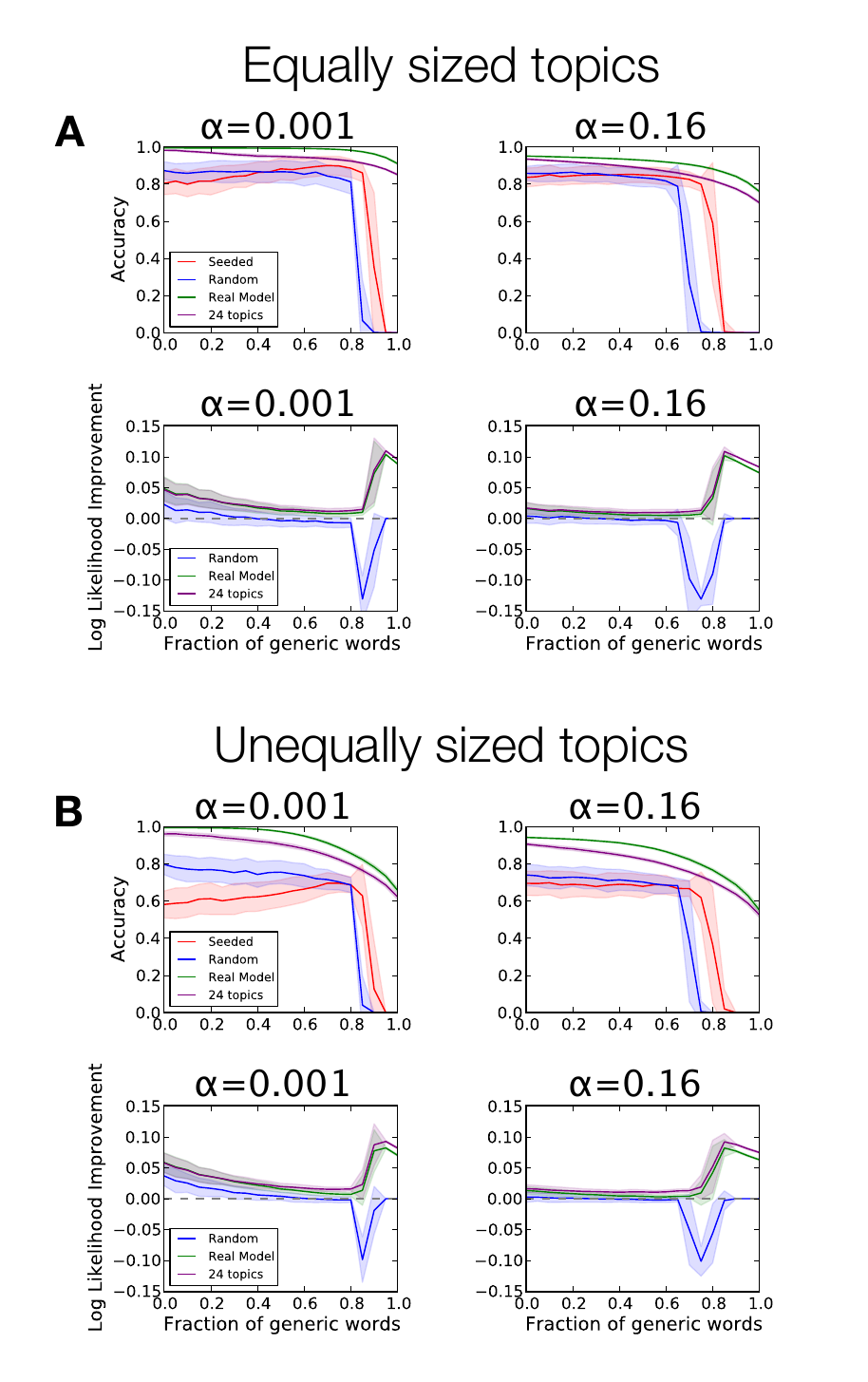}
\caption{How the initial conditions affect the performance of LDA. We checked four different ways of initializing the topics: \textit{random} and \textit{seeded} are the basic provided options. Real model refers to setting the underlying true parameters as initial conditions. $24$ \textit{topics} refers to the right initial conditions where we added $4$ small topics peaked on a single randomly chosen word. The log likelihood improvement is defined as the relative difference in the log likelihood we get with the different initial conditions compared to the \textit{seeded} initialization. The plot shows mean values and standard deviations.}
\label{initial_conditions}
\end{center}
\end{figure}

In this section, we discuss how the initial conditions affect the performance of LDA optimization. Two standard different ways of initializing the topics have been considered: \textit{random} and \textit{seeded}. The former assigns random initial conditions while the latter uses randomly sampled documents as seeds. We used both throughout the whole study, but we have only shown the \textit{seeded} version in the WoS dataset (the difference in performance is not appreciable, though). Here we compare these two initializations with the performance of the method when we guess the best possible initial conditions, meaning we start from the actual generative model (Fig.~\ref{initial_conditions}).

Similarly to the language test, starting from the generative model as initial conditions, we get an outstanding performance, which is also the optimal one in terms of likelihood. 
However, we checked that if we slightly change the number of topics, the performance gets worse and the likelihood improves. In Fig.~\ref{initial_conditions}, we show both performance and likelihood. $24$ \textit{topics} refers to a model close to the generative one, but where we added $4$ small topics, for which only one single word can be drawn: more precisely, we pick a word at random $w_r$ and we define these small topics with word probability distributions $p(word|topic)=\delta_{word,w_r}$. LDA will grow these small topics to increase the likelihood, overfitting the data and getting a worse performance.
This is the main reason why we decided to threshold small topics in the Web of Science dataset (see Sec.~\ref{network_sec}).

\subsection{TopicMapping guess}

\begin{figure}[h!]
\begin{center}
\includegraphics[width=8cm]{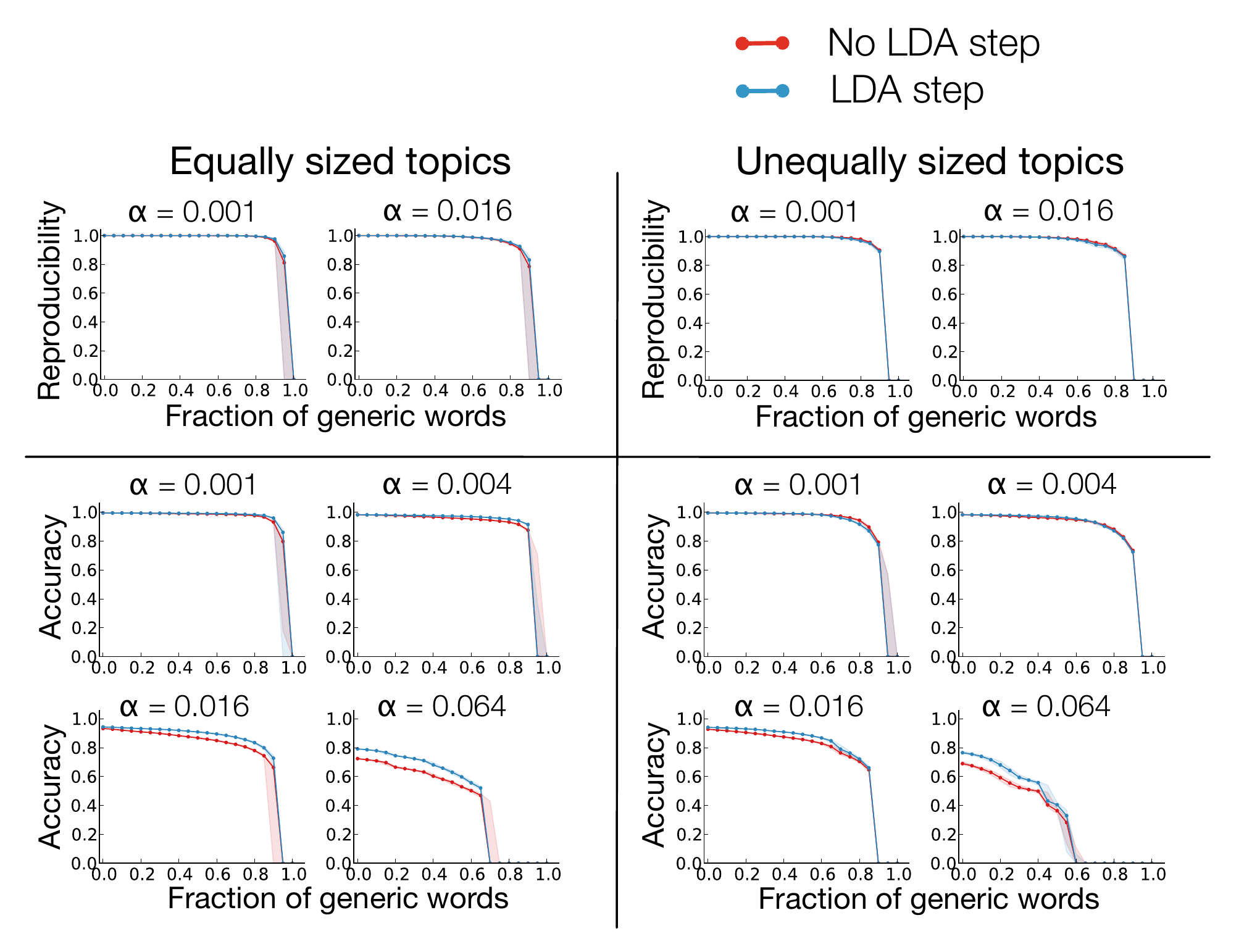}
\caption{Performance of TopicMapping on the synthetic datasets, before and after running LDA.}
\label{justplsa}
\end{center}
\end{figure}

Here, we show the performance of TopicMapping just for the guess, i.e. before running the LDA optimization (see Fig.~\ref{justplsa}).
We do not show the results for the language test because, in that case, there is no difference at all. In the systematic tests, instead, running LDA as a last step slightly improves the performance of the algorithm, although the difference is not dramatic.
We found a remarkable difference only in the Wikipedia dataset (see Sec.~\ref{wikipedia}), where the topic distribution provided by the guess was highly heterogeneous.

\section{Asymmetric LDA}
\label{asym_LDA}

In this section we discuss the results we obtain using asymmetric LDA \cite{wallach2009rethinking} (\url{http://mallet.cs.umass.edu}). 
The algorithm has two main differences respect with the other LDA method we used throughout the study: first, the prior probabilities of using a certain topic are not all equal, and, second, the optimization algorithm is based on Gibbs-sampling rather than variational inference \cite{blei2003LDA}.

\begin{figure}[h]
\begin{center}
\includegraphics[width=8cm]{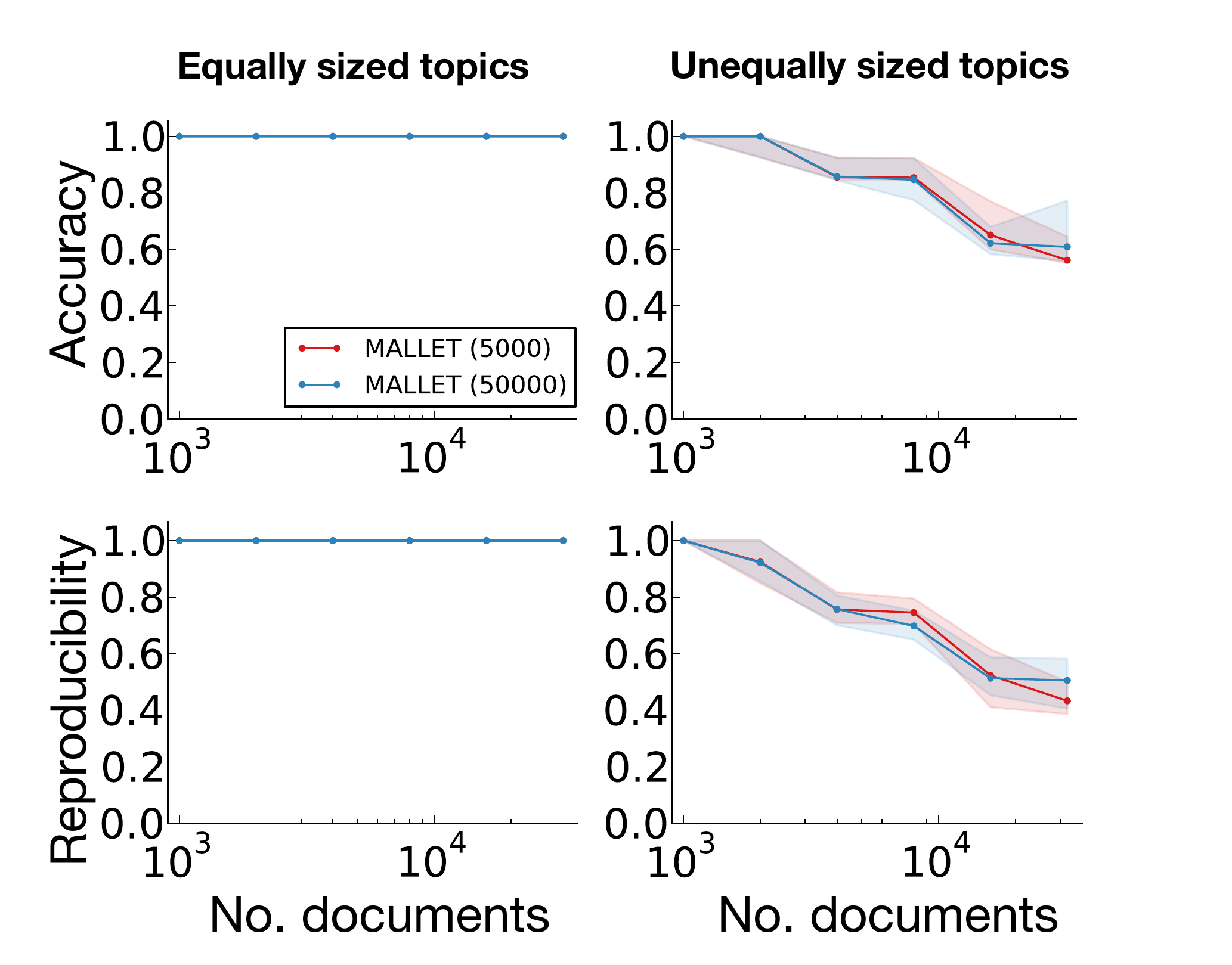}
\caption{Performance of asymmetric LDA in the language test (same as Fig.~2 in the main text). We used $5,000$ and $50,000$ iteration for Gibbs sampling and we input the correct number of languages in the algorithm. We optimize the hyper parameters each 100 iterations but performance is barely affected by the optimization interval. Curves are the median values and the shaded areas indicate 25th and 75th percentiles.}
\label{asym_fig1}
\end{center}
\end{figure}

\begin{figure}[h]
\begin{center}
\includegraphics[width=8cm]{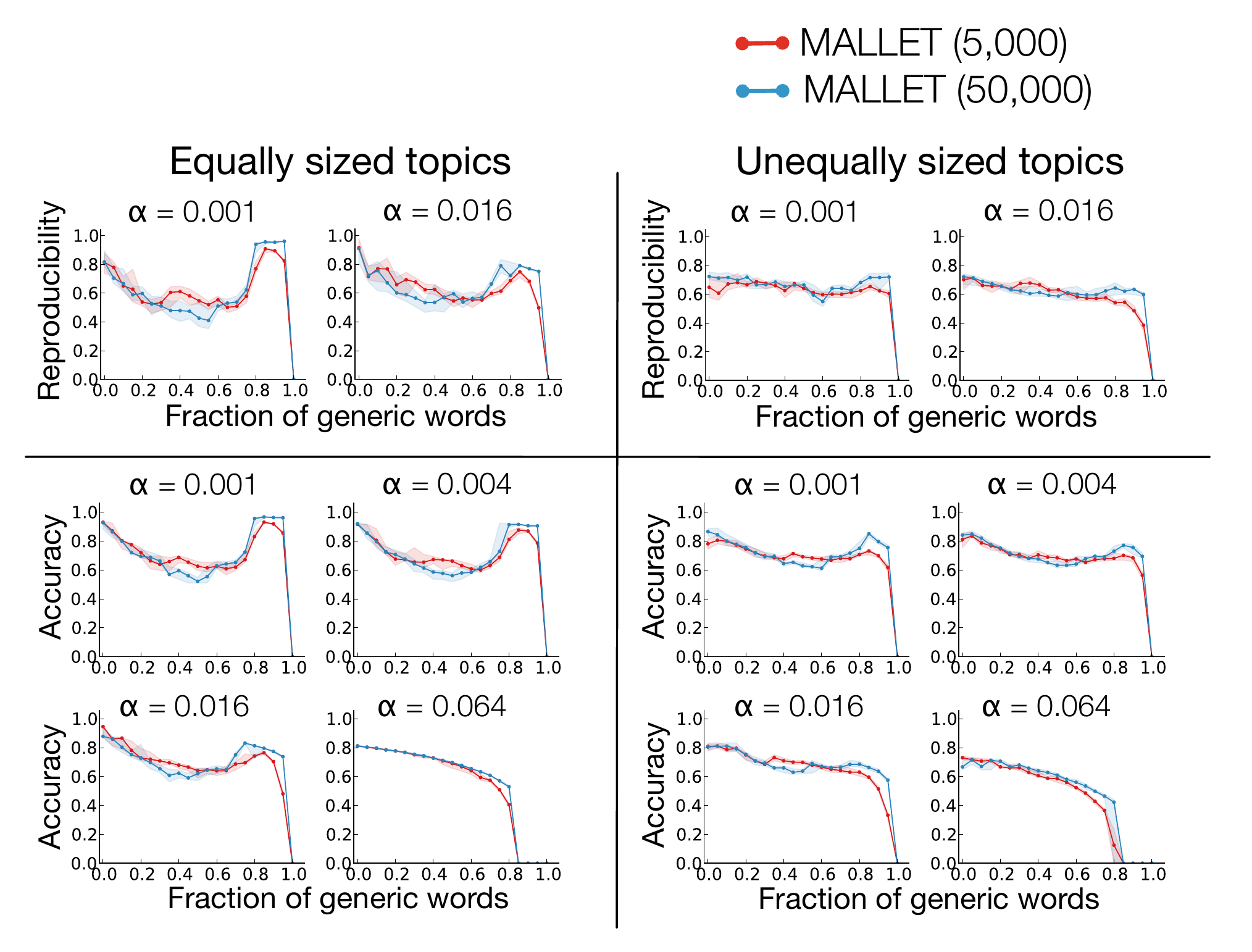}
\caption{Performance of asymmetric LDA on the tests presented in Fig.~4 in the main paper. Curves are median values and colored areas are 25th and 75th percentiles.}
\label{asym_fig2}
\end{center}
\end{figure}

Fig.~\ref{asym_fig1} shows that the algorithm performs better than symmetric LDA in the language test, although it still struggles recognizing the languages if the number of documents is large and the language probabilities are  unequal. The performance on the synthetic graphs is better to standard LDA, (see Fig.~\ref{asym_fig2}) for certain parameters only.

\section{The hierarchy of WoS dataset}
\label{wos_hierarchy}

In this section, we study the subtopic structure of the Web of Science dataset. In fact, we expect to  find subtopics in each journal. Although we do not know any ``real" topic model to compare with, we can still measure the reproducibility of the algorithm.

\begin{figure}[h]
\begin{center}
\includegraphics[width=8cm]{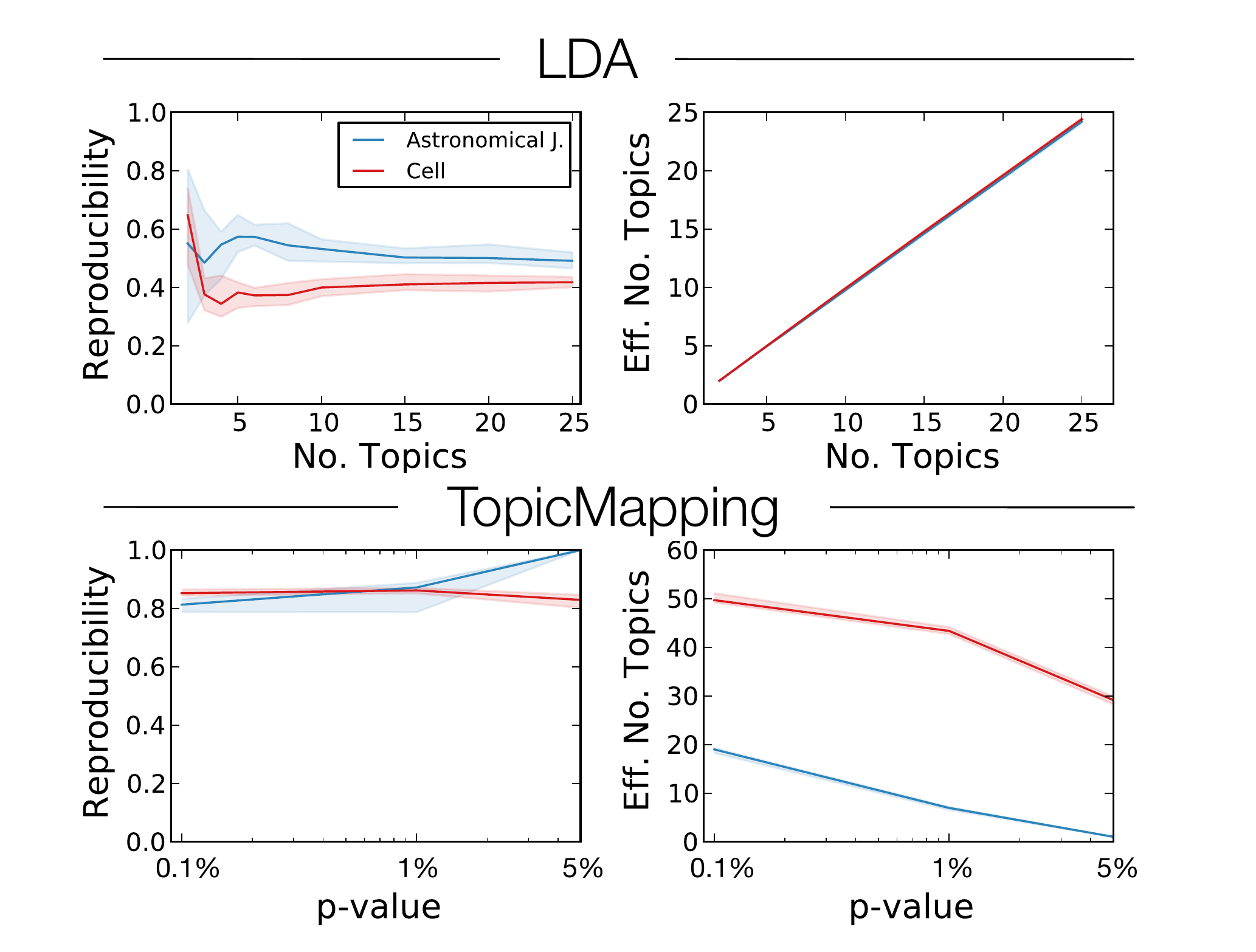}
\caption{Reproducibility and effective number of topics for LDA and TopicMapping for the scientific abstracts of Astronomical Journal and Cell. The number of topics can be tuned in LDA changing the input number of topics. Similarly, in TopicMapping the resolution can be tuned to some extent  filtering words with different $p$-values. However, this effect is present only in corpora with a less defined topic structures than the language test or the  synthetic graphs, for instance. Median and 25th and 75th percentiles are shown.}
\label{wos_subtopcs_lda}
\end{center}
\end{figure}

Similarly to what we observed above, we find again that standard LDA is not reproducible and the effective number of topics is strongly affected by the input number of topics, see Fig.~\ref{wos_subtopcs_lda}.

For TopicMapping, we observe that the number of topics is affected by the $p$-value we choose for filtering the noisy words. 
This is not what happens in all the other tests we have presented so far, which have a rather clear topic structure: therefore, choosing a $p$-value of $5\%$ or $1\%$ barely makes any difference. 
Instead, in analyzing Astronomical Journal abstracts, for instance, the topic structure is not so sharp anymore and we do observe that reducing the $p$-value provides a higher number of topics. Fig.~\ref{wos_subtopcs_lda} shows the results. For Astronomical Journal, with a $p$-value of $5\%$ we only observe one topic. Decreasing the $p$-value to $1\%$ we start observing sub-topics like: ``galaxi* observ* emiss*", ``star cluster metal"  or ``orbit system planet". For Cell, we also observe that the effective number of topics increases for smaller $p$-values. However, in both cases, TopicMapping is much more reproducible.

\section{Computational complexity}
\label{sec_time}

For a given vocabulary size, LDA's complexity is proportional to the number of documents times the number of topics.

\begin{figure*}
\begin{center}
\includegraphics[width=6.5in]{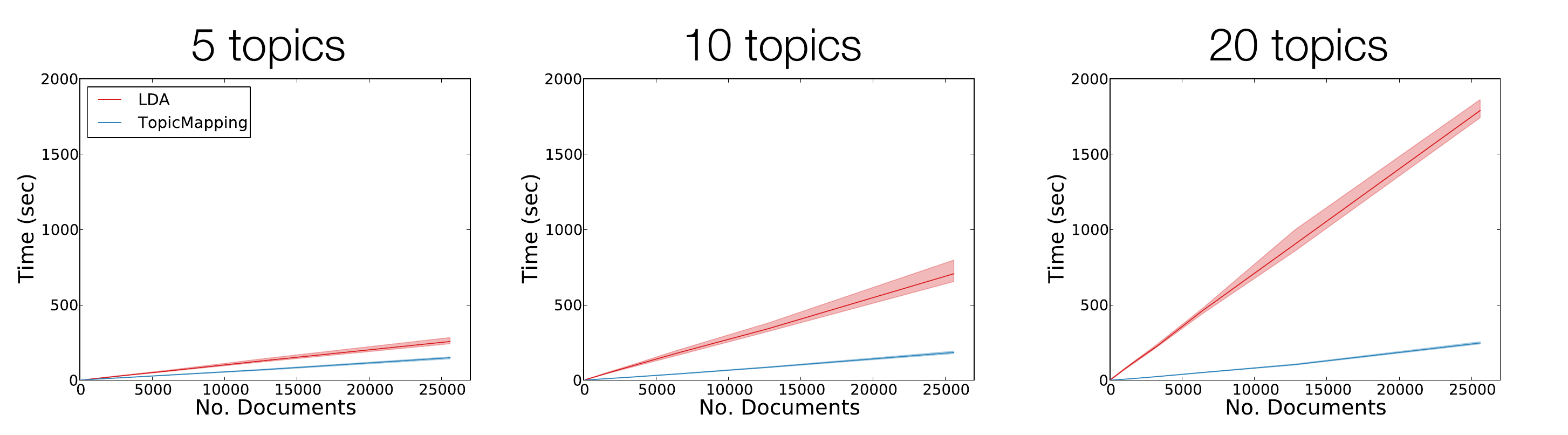}
\caption{Time needed for the execution of standard LDA and TopicMapping (before the LDA step) on synthetic corpora. Similarly to the other tests, we used a fixed vocabulary of 2000 unique words and 50 words per document. We set $\alpha=10^{-3}$ and the generic words are $30\%$. Both algorithms' complexity is linear in the number of documents. However, TopicMapping can be significantly faster if the number of topics is large.}
\label{time_complexity}
\end{center}
\end{figure*}

The computational complexity of TopicMapping's guess is also linear with the number of documents.
In particular, building the graph costs $O(\sum_d u_d^2)$, where $u_d$ is the number of unique words in document $d$. Infomap's complexity is of the same order of magnitude (smaller if we filter links), because the algorithm runs in a time proportional to the number of edges in the graph. Local PLSA-likelihood optimization is also linear in the number of documents, and can scale better than LDA with the number of topics, if the assignments of words to topics is sparse. In fact, we use sparse data structures to compute the topics for each document and each word, meaning that for each document, for instance, we do not handle a list of all topics (including never used topics), but only a list of the topics the document actually makes use of. Indeed, this enables the algorithm to scale much better with the number of topics (see Fig.~\ref{time_complexity}) on the synthetic datasets.

As a further example, to analyze the WoS corpus, TopicMapping takes $\sim 25$ minutes on a standard desktop computer.  LDA takes $\sim 20$ minutes for finding models with 6 topics and $~120$ minutes for models with 24 topics.

\newpage

\section{Topics in Wikipedia}
\label{wikipedia}

We have collected a large sample of the English Wikipedia (May 2013). The whole datasets comprises more than 4 million articles. However, since most of them are very short articles (stubs),  we decided to consider only articles with at least 5 in-links, 5 out-links and 100 unique words. Also very specific words (such as those which appear in less than 100 articles) have been pruned. This gives us a dataset of 1,294,860 articles 118,599 unique words and $\sim800$ millions words in total.

In order to get results quickly, we decided to parallelize most of the code. For building the network we used 9 threads,
each one was assigned a fraction of the total word pairs we had to consider. Doing so, we were able to construct the graph of words in roughly 12 hours. Infomap is extremely fast: each run of the algorithm takes about one hour and we ran it 10 times. After that, we ran the filtering  algorithm with a single thread, taking less than one day (we set a filtering step of 0.05). Finally, we parallelized the LDA optimization on about 50 threads: doing so, each iteration took about an hour.

\begin{figure*}
\begin{center}
\includegraphics[width=7in]{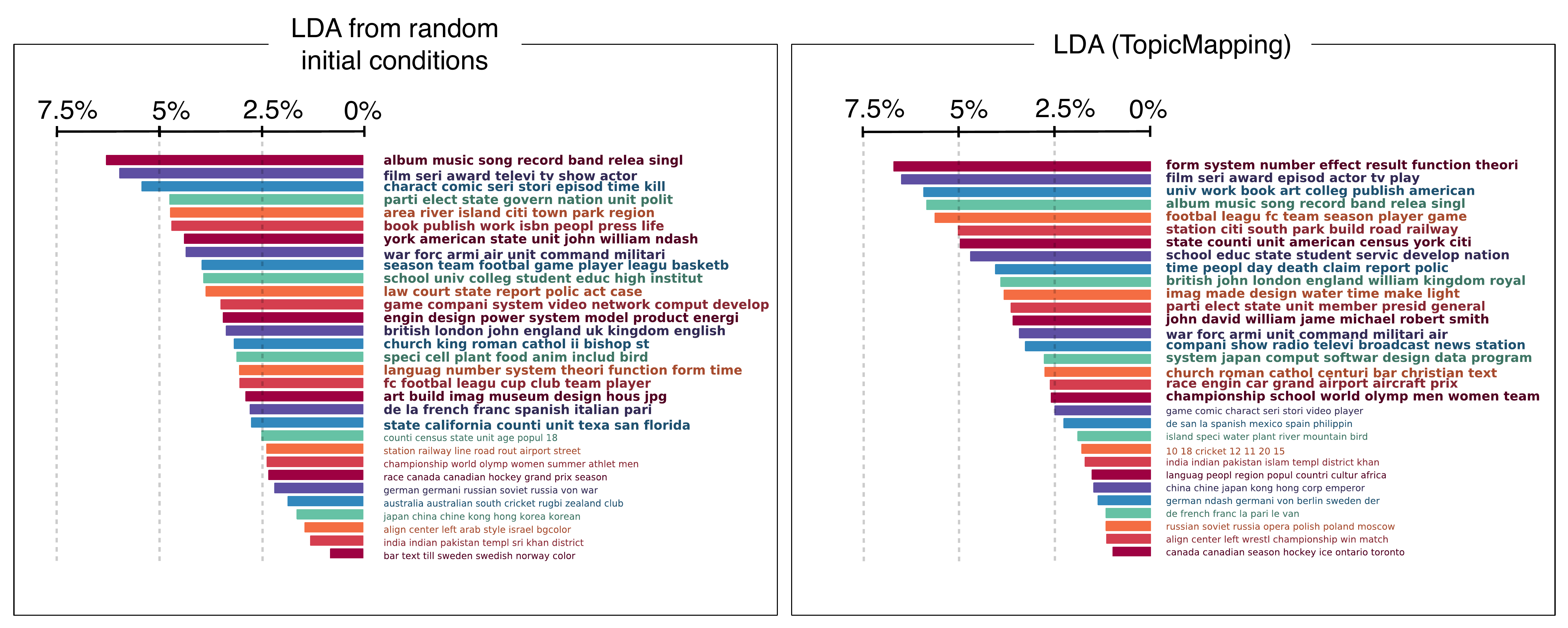}
\caption{Topic found in a large Wikipedia sample by standard LDA and TopicMapping with full LDA optimization. For each topic, we show the top 7 words. Bold fonts are used for the top topics which account for $80\%$ of the total.}
\label{fig::wiki}
\end{center}
\end{figure*}

In the main paper, we have shown the results of TopicMapping after running LDA optimization for one single iteration. The inset was obtained running the algorithm on the sub-corpus consisting of all words which were more likely drawn from the first topic.  
Fig.~\ref{fig::wiki}, instead, shows the results after the full LDA optimization. For comparison, we also show the results starting the algorithm with random initial conditions.
Interestingly, in this dataset, LDA optimization changed our guess significantly. This is not what happens in any of the other datasets we have tested, for which the topics in our guess were less heterogeneous (see Sec.~\ref{seb_sec::modelcomp}).

\section{TopicMapping as a likelihood optimization method}
\label{likelihood::sec}

Here we discuss to which extend TopicMapping provides models with better likelihood compared to standard LDA.
Indeed, in controlled test cases as the synthetic tests we have presented in this work, TopicMapping generally finds
better models in terms of likelihood and this explains why it performs better (the actual generative model has the highest likelihood).

\begin{figure}
\begin{center}
\includegraphics[width=8cm]{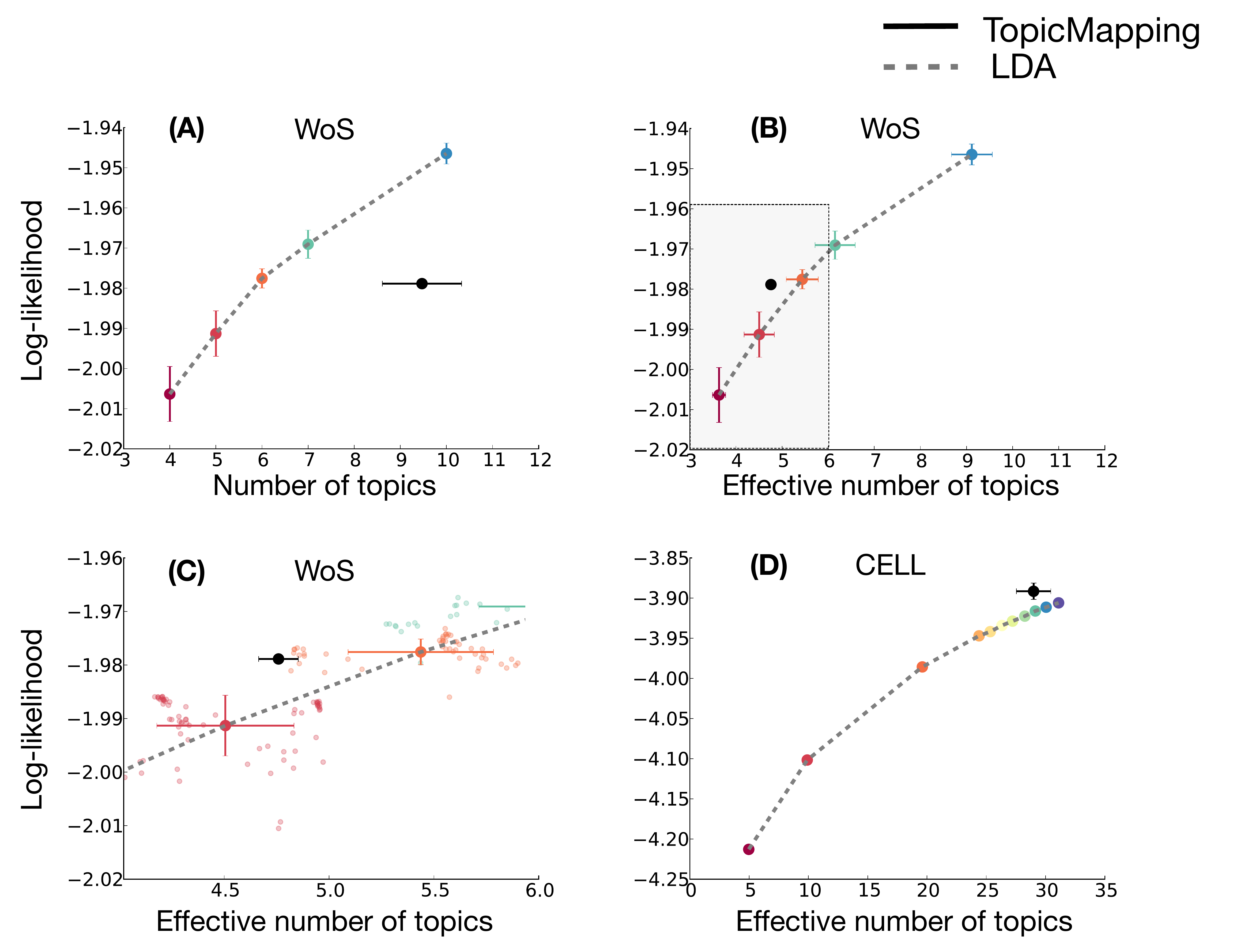}
\caption{Comparison between TopicMapping and standard LDA in terms of likelihood, for the Web of Science dataset we described in the main paper, and for Cell.  Colors represent the results from standard LDA with different number of topics as input. \textbf{A}. TopicMapping does not provide better likelihood if we compare models with the same number of topics. \textbf{B}. Comparison for Web of Science between model with the same effective number of topics. \textbf{C}. Zooming in the shaded area in \textbf{B}, we can see that TopicMapping performs better on average, although standard LDA is sometimes comparable. Indeed splitting Cell and merging Schizophrenia Bullettin and American Economic Review gives very comparable results in terms of likelihood. \textbf{D}.  Comparison for Cell.}
\label{fig::lik}
\end{center}
\end{figure}

In real cases, as we discussed in Sec.~\ref{seb_sec::modelcomp}, the likelihood can be maximized splitting large topics in subtopics and merging 
smaller topics. Therefore, if we compare the likelihood found by TopicMapping and the one found by variation inference \cite{blei2003LDA} as a function of the number of topics, TopicMapping does not provide models with higher likelihood. However, this comparison heavily penalizes TopicMapping, which often provides models with a broad distribution of topics, and many of them are barely used at all. 
We then argue that comparing models with the same number of effective topics is a more fair comparison. Doing so, Fig.~\ref{fig::lik} shows that, indeed, TopicMapping's models have often higher likelihood. However, the difference is not dramatic as we can see from the inset of Fig.~\ref{fig::lik}, because of the degeneracy of the likelihood landscape.

\section{Appendices}
\label{app_secs}

\subsection{Likelihood of English documents in the language test}
\label{appendixa}

In this section, we compute the likelihood of the alternative model for the English documents (Sec.~\ref{sec_deg}).
Let us call $a$ and $b$ the number of English words in the first group and in the second group respectively. We have that:

\begin{equation*}
a+b=N_w \,\,\, \,\,\, \,\,\, \,\,\, \textrm{and} \,\,\, \,\,\, \,\,\, \,\,\,  a f_1+ b g_1 =1,
\end{equation*}
and the equivalent holds for $f_2$ and $g_2$.
When we write an English document, we randomly sample words from the English vocabulary. This means that the probability that $n_a$ words fall in the first group, and $n_b=L_d-n_a$ in the second, follows a binomial distribution:

\begin{equation}
\label{binomial_eq}
p(n_a)=  \frac {L_d!} {n_a! (L_d-n_a)!} \Big (\frac{a}{N_w} \Big )^{n_a} (1-\frac{a}{N_w} \Big )^{L_d-n_a} .
\end{equation}

The last ingredient is how to decide which dialect a document should be fitted with. Let us define a threshold $T$ such that, if $n_a \geqslant T$ we use the first dialect, and we use the second otherwise. Without loss of generality, we also assume that $T \geqslant 1$, because otherwise we go back to the one single dialect case (Eq. \ref{likelihood1}).

Let us call  $\mathcal{L}^{'}_E$ the likelihood of an English document in this model. Its average can be written as:

\begin{equation}
\label{alt_likelihood}
\langle\log \mathcal{L}^{'}_E \rangle= \sum_{n_a=T}^{L_d} p(n_a) \log{\mathcal{L}^{'}_1(n_a)} + \sum_{n_a=0}^{T-1} p(n_a) \log{\mathcal{L}^{'}_2(n_a)} ,
\end{equation}

where

\begin{equation*}
 \log \mathcal{L}^{'}_1(n_a) = n_a \log{f_1} + (L_d - n_a) \log{g_1} ,
\end{equation*}

and the same equation holds for $ \mathcal{L}^{'}_2$ replacing $f_1$ with $f_2$ and $g_1$ with $g_2$.

We can compute the optimal values for $f_1$ and $f_2$ simply setting derivatives to zero:

\begin{equation*}
 \frac{\partial \log \mathcal{L}^{'} }{\partial f_1} = \sum_{n_a=T}^{L_d-1} p(n_a) \Big (  \frac{n_a}{f_1} + \frac{a (L_d - n_a)}{ a f_1 -1 }   \Big ) + p(L_d) \frac{L_d}{f_1}= 0 .
\end{equation*}

If we call:

\begin{equation*}
\omega_1 = \sum_{n_a=T}^{L_d} p(n_a) \textrm{,} \,\,\, \,\,\,  m_{a_1}  = \frac{\sum_{n_a=T}^{L_d} p(n_a) n_a}{\omega_1}  \textrm{,}
\end{equation*}
\begin{equation*}
p_{a_1}= \frac { m_{a_1} }{ L_d} \,\,\, \textrm{and} \,\,\,   \mu_a = L_d a / N_w ,
\end{equation*}

the optimal $f_1$ and $f_2$ can be written as:

\begin{equation*}
f_1= \frac{p_{a_1}}{a} \quad \textrm{and} \quad  f_2= \frac{p_{a_2}}{a} .
\end{equation*}

$\omega_1$ is how often we use the first dialect, $\mu_a$ is the expected number of words which fall in the first group of English words, $p_{a_1}$ is the probability of using words from the first group, given that we are using the first dialect. We also have that:

\begin{equation*}
\omega_2= 1 - \omega_1 \quad \textrm{and} \quad \omega_1 p_{a_1} + \omega_2 p_{a_2} = p_a = \frac{a}{N_w} .
\end{equation*}

For group $b$, we have:

\begin{equation*}
p_{b_1} = 1- p_{a_1} \quad \textrm{and} \quad p_{b_2} = 1- p_{a_2}.
\end{equation*}

We can now compute the expected log-likelihood of Eq. \ref{alt_likelihood}:

\begin{equation*}
\frac{\langle\log \mathcal{L}^{'}_E \rangle}{L_d}  =   \omega_1 [ p_{a_1} \log{\frac{p_{a_1}}{a}}  +  p_{b_1} \log {\frac{p_{b_1}}{b}} ] +
\end{equation*}
\begin{equation*}
\omega_2 [ p_{a_2} \log{\frac{p_{a_2}}{a}}  +  p_{b_2} \log {\frac{p_{b_2}}{b}} ] .
\end{equation*}

Calling the entropy of a binary variable $h(p)= - p\log{p} - (1-p) \log{(1-p)}$,  we get:

\begin{equation}
\label{lik_diff_eq}
\frac{\langle\log \mathcal{L}^{'}_E \rangle - \mathcal{L}^{'}_{\textrm{true}}  }{L_d}  =  - \omega_1 h(p_{a_1}) - \omega_2  h(p_{a_2}) + h(p_a) .
\end{equation}

Now, the problem is to find, for given $L_d$ and $N_w$, which choice of the parameters $a$ and $T$ maximizes Eq. \ref{lik_diff_eq}.
It turns out that there are two different regimes depending on the condition $N_w \geqslant L_d$.

If $N_w \geqslant L_d$, a possible strategy is to assume $T=1$. This means that we use the second dialect if (and only if) there are no words in the first group. This means that $p_{a_2}=0$.

In fact, using the equations above, we get:

\begin{equation*}
\omega_1= 1- \Big (1- \frac{a}{N_w} \Big)^{L_d}  \,\,\, p_{a_1}= \frac{a}{N_w \omega_1}  \,\,\, \textrm{and} \,\,\, p_{a_2}= 0 .
\end{equation*}

It is possible to prove that for $L_d \gg 1$ and $N_w \geqslant L_d$ the maximum is attained for:

\begin{equation*}
\omega_1=\frac{1}{2} \quad \textrm{and} \quad p_a=\frac{\log2}{L_d} ,
\end{equation*}

and disregarding size effect due to $a$ being an integer:

\begin{equation*}
\langle\log \mathcal{L}^{'}_E \rangle - \mathcal{L}^{'}_{\textrm{true}} \simeq (\log{2})^2 \simeq 0.4804.
\end{equation*}

In the second case, $L_d \geqslant N_w$, we restrict ourselves to considering $T=L_d / N_w$. In the limit $L_d \gg 1$,  using the Gaussian approximation of the binomial distribution in Eq. \ref{binomial_eq}, we get:

\begin{equation*}
\omega_1 = \frac{1}{2} \quad \textrm{and} \quad p_a \simeq \frac{a}{N_w} + \sqrt{\frac{2 a }{\pi N_w L_d}},
\end{equation*}

\begin{equation*}
\langle\log \mathcal{L}^{'}_E \rangle - \mathcal{L}^{'}_{\textrm{true}}  \simeq \frac{N_w}{\pi (N_w -a)}.
\end{equation*}

If also $N_w \gg 1$, the difference is independent of $a$:

\begin{equation*}
\langle\log \mathcal{L}^{'}_E \rangle - \mathcal{L}^{'}_{\textrm{true}}  \simeq \frac{1}{\pi} \simeq 0.3183.
\end{equation*}

In conclusion, the log-likelihood per document of the alternative model (given that we use English), is bigger that the one of the generative model, and, remarkably, the difference varies from roughly $0.5$ to $0.3$, so that it is  substantially independent of all the parameters of the model. Since we can divide the English words in two arbitrarily groups, we can actually have a large number of alternative models. For instance, if we have $L_d=100$, and $N_w=1000$, the model with highest likelihood splits English in two groups of $7$ and $993$ words, so that the number of alternative models becomes:

\begin{equation*}
\binom{1000}{7} \simeq 10^{17} ,
\end{equation*}

and there are many more alternative models with slightly smaller likelihood: for instance using $a=500$, the likelihood of the alternative model is $99.6 \%$ the likelihood we obtain for $a=7$, but  the number is $\simeq 10^{300}$. All these models are likely local maxima of the log-likelihood for Expectation-Maximization algorithms.

\subsection{Derivation of Eq.~\ref{modelcomeq}}
\label{deghier}

Let us start computing  $\log {\mathcal{L}^{'}_{\textrm{M1}}}$, the log-likelihood per document for the model where the subtopics are merged and all languages are recovered. We recall that the symbol $'$ means that the likelihood is computed given that we know the topics of the document.

If we merge the two English subtopics, the common words $(C)$ have probability $1/(U+C)$ while the words only used in one of the two subtopics $(2U)$ have probability $1/(2U+2C) $. Therefore, the average log-likelihood per English document in Model 1 is:

\begin{equation*}
\label{underfiteq}
\langle \log {\mathcal{L}^{'}_{E} } \rangle = - L_d \frac{C}{U+C} \log(U+C) -  L_d \frac{U}{U+C}  \log (2U+2C),
\end{equation*}

which can be re-written as:

\begin{equation*}
\langle \log {\mathcal{L}^{'}_{E} } \rangle = - L_d \log (U+C) -  L_d \frac{U}{U+C} \log 2.
\end{equation*}

The log-likelihood per non-English document is:

\begin{equation*}
\log {\mathcal{L}^{'}_{NE} } = - L_d \log N_w .
\end{equation*}

Instead merging two languages which are not English (Model 2), we get:

\begin{equation*}
\log {\mathcal{L}^{'}_{M} } = - L_d \log 2 N_w .
\end{equation*}

The difference in the average log-likelihood between the two models becomes (we recall that $p_k$ is the probability of any non-English language):

\begin{equation*}
\langle \log {\mathcal{L}^{'}_{\textrm{M1}}} \rangle - \langle \log {\mathcal{L}^{'}_{\textrm{M2}}} \rangle= - p_E   L_d \frac{U}{U+C} \log 2 + 2 p_k L_d \log 2 .
\end{equation*}

 Eq.~\ref{modelcomeq} follows from the equation above. For asymmetric LDA, we also have to consider the difference in the language entropies.
 Accounting for that, we get:

\begin{equation*}
\langle \log {\mathcal{L}_{\textrm{M1}}} \rangle - \langle \log {\mathcal{L}_{\textrm{M2}}} \rangle =
\end{equation*}
\begin{equation*}
- p_E  ( L_d \frac{U}{U+C} -1) \log 2 + 2 p_k (L_d-1) \log 2 .
\end{equation*}
 
Then, Model 1 has higher likelihood than Model 2 if:

\begin{equation*}
 2 p_k (L_d-1)  > p_E  ( L_d \frac{U}{U+C} -1).
\end{equation*}
 
The correction from Eq.~\ref{modelcomeq} is $O(L_d^{-1})$.

\subsection{The Dirichlet distribution}
\label{about_dirichlet_distribution}

The Dirichlet distribution is frequently used in Bayesian statistics since it is the conjugate prior of the multinomial distribution. The distribution is parameterized by $K$ values $\alpha_{topic}$, and the support of the function is the standard $(K-1)-$simplex, i.e. the set of vectors \textbf{\textit{x}} of dimension $K$ such that $\sum_i x_i=1$ and all $x_i \geqslant 0$.  Clearly,  \textbf{\textit{x}} can be interpreted as a probability distribution. Moreover $\langle x_i \rangle=\alpha_i / \sum_{topic} \alpha_{topic}$.

\begin{figure}
\begin{center}
\includegraphics[width=8cm]{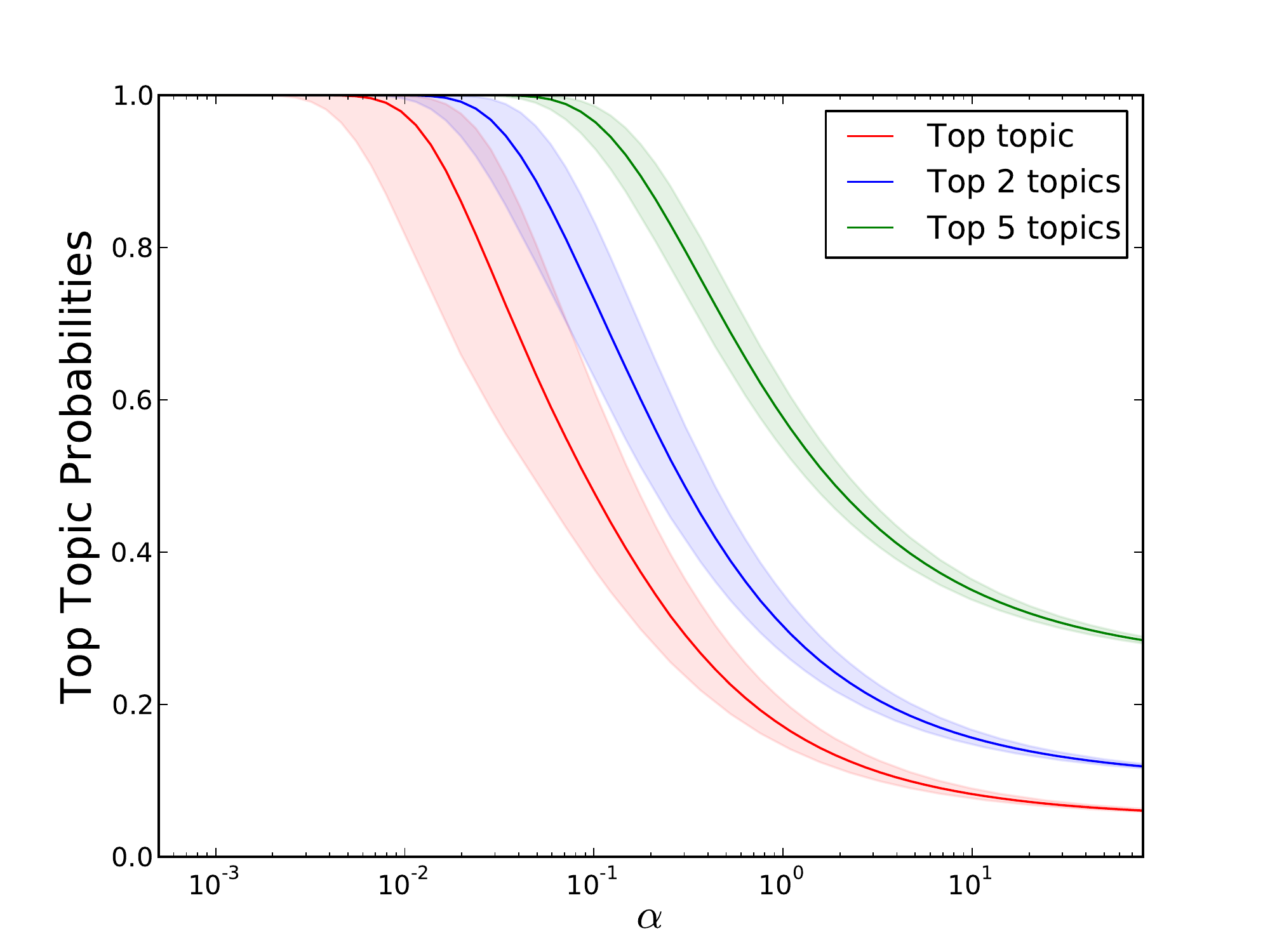}
\caption{For each document, we extract $p(topic|doc)$ from the Dirichlet distribution for several values of $\alpha$, setting $20$ equally probable topics. We then measure the probability of its most prominent topic (red curve) as well as the sum of the two and five largest topic probabilities (blue and green curves). The plot shows the median together with the $25\%$ and $75\%$ quantiles.  Small values of $\alpha$ lead to documents which are mostly assigned to one single topic: for instance, for $\alpha=10^{-3}$, the probability of the top topic is basically 1 and all others are zero. For  $\alpha=10^{-1}$, the top topic has roughly 0.5 probability, the second one has 0.2 and all the last 15 combined have less than 0.05. For large values like $\alpha \gtrsim10^2$, all topic probabilities tend towards equality, $p(topic|doc)=0.05$: this means that documents cannot be classified as they use all topics with equal probability. }
\label{dirichlet_fig}
\end{center}
\end{figure}

In generating the synthetic corpus, for each document, we use the same $\alpha_{topic} =  K \times p(topic) \times \alpha$ to draw $p(topic|doc)$ from  the Dirichlet distribution. In fact, even letting $\alpha$ depend on documents (but not on topics), this definition makes sure we get back the pre-defined topic probabilities, since $ \langle p(topic|doc) \rangle=p(topic)$.  Fig. \ref{dirichlet_fig} shows how $p(topic|doc)$ depends on $\alpha$ in the simple case of $20$ equiprobable topics.

\subsection{Measuring Perplexity}
\label{appendix_perple}

Perplexity is the conventional way to evaluate topic models' accuracy \cite{blei2003LDA}. Here, we briefly review how it is computed.

The spirit is to cross validate the model, whose parameters have been computed on a trained set of documents, looking at how well the model fits a small set of unseen documents. Therefore, the procedure is ($i$) to held out a fraction of documents from the corpus (typically $10\%$), ($ii$) train the algorithm using the remaining $90\%$ of the documents, ($iii$) infer the topic probabilities for the unseen documents $p(topic|doc)$ without changing the topics, i.e. $p(word|topic)$, ($iv$) compare the actual word frequencies $p(word|doc)$ of  the unseen documents with the topic mixture $q(word|doc)= \sum_{topic} p(topic|doc) \times p(word|topic)$.

\subsection{Algorithms' usage details}

For LDA, we used the implementation that can be found from \url{http://www.cs.princeton.edu/\~\ blei/lda-c/index.html}.
The stopping criterion in running LDA and PLSA was that the relative improvement of the log likelihood bound was less than $10^{-5}$ with respect to the previous iteration. In running LDA we let the algorithm optimize $\alpha$ as well. The initial value was set to  $\alpha=1$.

\bibliography{topic_modeling}

\end{document}